\documentclass[12pt]{article}
\usepackage{amsmath,amssymb}
\usepackage[left=2cm,right=2cm,top=2.5cm,bottom=2.5cm]{geometry}
\usepackage{float}
\usepackage{graphicx}
\graphicspath{{figures/}}
\usepackage{caption,subcaption}
\usepackage{listings}
\usepackage{url}
\usepackage{soul, color}
\usepackage{bm}
\usepackage{stackengine}
\usepackage{xcolor}
\usepackage[hidelinks]{hyperref}
\definecolor{olive}{rgb}{0.13, 0.55, 0.13}
\hypersetup{
    colorlinks=true,
    linkcolor=blue,
    filecolor=olive,      
    urlcolor=olive,
    citecolor=olive
}
\usepackage[export]{adjustbox}
\usepackage{verbatim}
\usepackage{siunitx}
\usepackage{stackengine}
\usepackage{comment}

\usepackage{pifont}

\usepackage{authblk}

\title{\bf Physics-Informed DeepONet Coupled with FEM for Convective Transport in Porous Media with Sharp Gaussian Sources\\} 
\author[1]{Erdi Kara,}
\author[2]{Panos Stinis}

\affil[1]{Department of Mathematics, Spelman College, Atlanta, GA, USA}
\affil[2]{Advanced Computing, Mathematics and Data Division,\\
Pacific Northwest National Laboratory, Richland, WA, USA}

\date{}           
\begin{document}
\maketitle
\begin{abstract}
We present a hybrid framework that couples finite element methods (FEM) with physics-informed DeepONet to model fluid transport in porous media from sharp, localized Gaussian sources. The governing system consists of a steady-state Darcy flow equation and a time-dependent convection–diffusion equation. Our approach solves the Darcy system using FEM and transfers the resulting velocity field to a physics-informed DeepONet, which learns the mapping from source functions to solute concentration profiles. This modular strategy preserves FEM-level accuracy in the flow field while enabling fast inference for transport dynamics. To handle steep gradients induced by sharp sources, we introduce an adaptive sampling strategy for trunk collocation points. Numerical experiments demonstrate that our method are in good agreement with the reference solutions while offering orders-of-magnitude speedups over traditional solvers making it suitable for practical applications in relevant scenarios.  Implementation of our proposed method is available in our
Github repository at \url{https://github.com/erkara/fem-pi-deeponet}.
\end{abstract}

\section{Introduction}

Fluid flow and solute transport in porous media play a critical role in many scientific and engineering applications, particularly in areas such as subsurface energy storage, carbon sequestration, geothermal systems, and groundwater contamination \cite{yang2024data, lyu2025multiscale, hesse2008mathematical, al2021comprehensive,sharma2020review,baqer2022review}. These processes are often modeled using coupled systems of partial differential equations (PDEs) that describe fluid pressure, velocity, solute concentration and other variables of interest over time and space. In practical scenarios, these governing equations depend on many parameters, including source characteristics, medium properties, and boundary conditions, which can vary significantly across applications \cite{helmig1997multiphase,chen2006computational}.

Classical numerical methods such as FEM has been the gold standard in the qualitative and numerical study of the underlying governing PDEs \cite{chen2006computational,lewis2004fundamentals,zienkiewicz2015finite}. By construction, these PDEs often naturally involve several parameters such as problem geometry, initial and boundary conditions as well as physical parameters representing permeability, porosity, diffusivity and so on. One of the fundamental limitation of conventional methods is that they usually solve one instance of the underlying PDE, thus limiting their practical use in applications requiring fast inference since every parameter configuration requires to run the solver again.

In recent years, physics-informed machine learning has emerged as a promising avenue to overcome these limitations by constructing fast surrogate models for PDEs. Physics-Informed Neural Networks (PINNs) were introduced as deep learning frameworks that embed physical laws (e.g. underlying PDEs) into the training objective \cite{karniadakis2021physics}. By penalizing the PDE residual in the loss function, PINNs can learn the solution of a given differential equation without requiring extensive labeled data, yielding a surrogate that obeys the underlying physics. This approach has been successfully applied to forward and inverse problems for fluid flow and other nonlinear PDE systems \cite{meng2025physics,cai2021physics}. However, a standard PINN is typically trained for one specific problem setup at a time (with fixed domain and parameters), which can limit its reusability across varying inputs.

To address this limitation, neural operator learning has emerged as a powerful alternative, enabling the learning of solution mappings across varying domain geometries, input parameters, and boundary conditions. Unlike PINNs, which approximate a single solution, neural operators learn mappings between function spaces, effectively capturing the solution operator of an entire parametric family of PDEs \cite{rosofsky2023applications,kovachki2023neural,lu2022comprehensive}. Although they require upfront training, neural operators allow fast inference across diverse scenarios without repeatedly solving the PDE from scratch, with ideally minimal loss of accuracy. 

Among these approaches, one of the most well-known and successful one is DeepOnet framework Lu et al \cite{lu2021learning}. DeepONet uses a dual-network architecture (with a “branch” net encoding input functions, and a “trunk” net encoding the output coordinates) to approximate nonlinear operators, supported by a universal approximation theorem for operators \cite{chen1995universal}. This enables learning an entire family of PDE solutions: for instance, mapping different source terms or boundary conditions to the corresponding solution fields \cite{jin2022mionet,kovachki2023neural}. In fact, these surrogates have been reported to produce solutions up to 1000x faster than standard FEM-based PDE solvers while maintaining good accuracy. Literature is rich with several extensions of the original DeepOnet, such Bayesian DeepOnet \cite{garg2023vb}, multi-fidelity DeepOnet \cite{howard2023multifidelity}, multi input-output DeepOnet \cite{jin2022mionet}.

In its original form, DeepONet is a supervised machine learning model and therefore requires input–output pairs. This implies that one must obtain such solutions from the underlying PDEs. Even for moderately sized problems, this involves collecting a large amount of high-fidelity data from numerical solvers, which can be very expensive. To address these challenges, Wang et al. incorporated physical constraints into the training process, giving rise to \textit{physics-informed DeepONets (PI-DeepONets)}, which use PDE residuals instead of supervised solution labels \cite{wang2021learning}. The fundamental idea relies on the fact that DeepONet outputs are differentiable with respect to input parameters (i.e., trunk and branch inputs), allowing for PINN-style physics enforcement in DeepONet via automatic differentiation \cite{baydin2018automatic}. The implication is that the network can learn operator mappings without requiring paired input–output datasets. The same team also introduced a modified DeepONet architecture designed to mitigate vanishing gradient issues and improve convergence in physics-informed operator learning via NTK theory \cite{wang2022improved}. In this architecture, authors also augmented the conventional branch and trunk networks of original DeepOnet with encoder layers that embed the input function and evaluation coordinates, respectively, into higher-dimensional latent spaces. These embeddings are then injected into each hidden layer via pointwise operations, enabling hierarchical interaction between the networks throughout the forward pass. This architectural modification facilitates more stable signal propagation and has been empirically shown to outperform conventional DeepONets, particularly in the physics-informed setting without paired training data. This work forms the foundation of our work in this paper that will be detailed in the subsequent sections.

In the study of porous media flow, several data-driven studies have been proposed in recent years. A recent review by Yang et al. provides a broad overview of such methods \cite{yang2024data}. In the context of physics-guided machine learning, there is a relatively rich literature applying PINNs to porous media problems, motivated by the need for fast surrogates in real-world applications.

Tartakovsky et al. developed a PINN framework for inferring spatially varying parameters and unknown constitutive relations in subsurface flow, demonstrating accurate recovery from sparse data under Darcy and Richards models \cite{tartakovsky2020physics}. Jagtap and Karniadakis introduced XPINNs, a space–time domain decomposition approach where separate PINNs are deployed in irregular subdomains to improve parallelism and flexibility for complex PDEs \cite{jagtap2020extended}. Yan et al. proposed a gradient-based PINN (GDNN) that regularizes training through physically motivated differential operators; applied to CO$_2$ injection scenarios, the model captures nonlinear multiphase flow and extends naturally to convolutional architectures \cite{yan2022gradient}. Zhang et al. designed a physics-informed deep convolutional network (PIDCNN) for transient Darcy flow, combining finite volume loss enforcement with a well model to capture steep pressure gradients \cite{zhang2022physics}. Cai et al. developed a PINN framework for Biot’s poroelasticity equations using a fixed-stress splitting strategy to decouple pressure and displacement, with Darcy flow as the governing mechanism for fluid motion \cite{cai2023combination}. For additional PINN applications in porous media, see also \cite{amini2023inverse,kadeethum2020physics,kashefi2022physics}.

Compared to PINNs, operator learning studies in porous media are more limited. Huang et al. introduced the \textit{Porous-DeepONet} framework, which augments DeepONet with convolutional networks to encode image-based porous structures \cite{huang2024porous}. They also proposed Porous-PI-DeepONet for physics-informed training and Porous-DeepM\&Mnet for multi-physics coupling, demonstrating strong generalization across various reactive transport problems. Yan et al. employed Fourier Neural Operators to model multiphase flow under heterogeneous permeability, showing accurate predictions and speedups in CO$_2$ sequestration scenarios \cite{badawi2025neural}. Kumar et al. introduced Multi-task DeepONet (MT-DeepONet), which solves diverse PDEs across varying source terms and geometries using a modified branch network and binary masking \cite{kumar2025synergistic}. Mandl et al. proposed Physics-Informed Separable DeepONet (PI-Sep-DeepONet), which factorizes the trunk network for efficient operator learning in high-dimensional problems, and applied it to Biot’s consolidation system, a Darcy-type flow setting without transport \cite{mandl2025separable}.

In this paper, we study the problem of fluid infusion into a porous medium from a sharp, localized Gaussian source. The governing model consists of three coupled PDEs: a steady-state Darcy system that determines the pressure and velocity fields, and a time-dependent convection–diffusion equation that models the transport of the infused fluid. Our objective is to learn the mapping from the source function to the resulting concentration profile. While operator learning has been applied to porous media problems in various forms, our work introduces several key innovations.

First, we propose a modular solution strategy in which the Darcy flow problem is solved using a finite element method (FEM), and only the transport component is learned using a physics-informed DeepONet. Unlike most existing approaches that treat the entire PDE system as a single learning task, we couple high-fidelity numerical solvers and operator learning in a sequential and physically consistent way. This allows us to preserve FEM-level accuracy in the velocity field while benefiting from the fast inference and generalization properties of DeepONet in the transport stage. Notably, despite the growing interest in operator learning, the use of PI-DeepONet, particularly in porous media applications, remains extremely rare in the current literature.

Second, to handle the sharp spatial localization of the Gaussian sources, we introduce a simple yet effective adaptive sampling strategy for trunk collocation points. This improves the resolution of steep gradients in the solution, which would otherwise challenge standard PINN or DeepONet training pipelines. To the best of our knowledge, this work is one of the few that applies physics-informed DeepONet to porous media transport problems and the first to combine it in a tightly coupled manner with FEM solvers.

\section{Problem Description}

\subsection{Convective Fluid Transport in Porous Media}

In this section, we describe the coupled systems of partial differential equations(PDE) to model the delivery and transport of an externally infused fluid carrying a dissolved solute into a porous medium. The system consists of two components: a steady-state Darcy flow and mass balance model that determines the pressure and velocity fields in the porous medium, and a time-dependent convection–diffusion equation that governs the evolution of the solute concentration.

\begin{subequations} 
\label{eq:pdesys}
\begin{align}
\mathbf{v} &= -\frac{\mathbf{K}}{\mu} \nabla p && \text{in } \Omega, \label{eq:darcy} \\
\nabla \cdot \mathbf{v} &= -\alpha p + \beta_1 f(x) && \text{in } \Omega, \label{eq:div} \\
\frac{\partial c}{\partial t} &= \nabla \cdot (D \nabla c) - \nabla \cdot (\mathbf{v} c) + \beta_2 f(x) && \text{in } \Omega, \label{eq:transport}
\end{align}
\end{subequations}

where $\mathbf{v}$ is the Darcy velocity, $p$ is the fluid pressure, and $c$ is the solute concentration. Here, $\mathbf{K}$ is the symmetric, positive-definite permeability tensor of the medium, $\mu$ is the dynamic viscosity, $D$ is the diffusion tensor while $f(x)$ is the spatially distributed source intensity, scaled by $\beta_1$ and $\beta_2$ for the fluid and solute, respectively.

We impose the following boundary conditions on the domain boundary $\partial \Omega$:
\begin{subequations} 
\begin{align}
p &= 0, \quad \mathbf{v} \cdot \mathbf{n} = 0 && \text{on } \partial \Omega, \label{eq:pressure_vel}\\
\mathbf{n} \cdot \nabla c &= 0 && \text{on } \partial \Omega, \label{eq:neuman} 
\end{align}
\end{subequations}
where $\mathbf{n}$ denotes the outward unit normal vector.

The first equation in the system \eqref{eq:darcy} is Darcy’s law that models fluid motion at low Reynolds number, where viscous forces dominate and inertial effects can be neglected. Darcy’s law expresses that the fluid flux is proportional to the pressure gradient, modulated by the local permeability and viscosity. Second equation \eqref{eq:div} expresses the local mass conservation which accounts for two effects: fluid loss through pressure-dependent sinks, such as leakage or matrix absorption ($-\alpha p$), and external fluid infusion through the source term $\beta_1 f(x)$. While the sink term is general to porous media flow, we note that a similar formulation appears in biology, where Starling’s law describes fluid exchange across semipermeable membranes as a balance of hydraulic and osmotic forces. Together, these equations form the system, which provides the stationary flow field and pressure distribution induced by the external infusion and internal leakage mechanisms.

The third equation \eqref{eq:transport}, which is of the standard convection-diffusion form, governs the transport of the dissolved solute. The first term on its right-hand side models diffusive or dispersive spreading of the solute, while the second term accounts for its advective transport by the fluid flow $\mathbf{v}$. The source term $\beta_2 f(x)$ models direct injection of solute into the domain, coinciding spatially with the fluid infusion.

The boundary conditions are chosen to reflect physically meaningful constraints. The zero-pressure condition, $ p = 0 $, assumes that the pressure has equilibrated at the domain boundary, meaning that the infused fluid has relaxed back to its environmental baseline pressure at the outer edge of the modeled region. This models an open system where pressure disturbances dissipate away from the injection site, and no artificial buildup of pressure occurs at the boundary. The no-flux condition on the velocity, $ \mathbf{v} \cdot \mathbf{n} = 0 $, reflects the presence of impermeable boundaries that do not permit net fluid flow across the boundary, ensuring that the fluid exchange is entirely governed by the internal sources and sinks. For the solute, the homogeneous Neumann condition, $ \mathbf{n} \cdot \nabla c = 0 $, enforces zero diffusive flux across the boundary, preventing solute from escaping the domain or being artificially introduced through the boundary.

In this work, we are primarily interested in fluid infusion driven by a localized source. To model this, we represent the source term $ f(\mathbf{x})$ as a normalized Gaussian profile centered at a prescribed location $\mathbf{x_0}$

\begin{align}
\label{eq:gauss}
f(\mathbf{x}) = \frac{1}{\mathcal{N}} \exp\left(-\frac{\lVert \mathbf{x} - \mathbf{x}_0 \rVert^2}{2\sigma^2}\right), \quad \mathcal{N} = \int_{\Omega} \exp\left(-\frac{\lVert \mathbf{x} - \mathbf{x}_0 \rVert^2}{2\sigma^2}\right) d\mathbf{x}.
\end{align}

where $ \sigma $ characterizes the spatial width of the source and determines how sharply localized the injection is around $\mathbf{x_0}$. This formulation is relevant in many practical applications where infusion occurs through a focused delivery mechanism, such as a point source in contaminant transport modeling \cite{ermak1977analytical,patrick2005contaminant,sun2014mathematical}, hydraulic fracturing \cite{tanikella2023dynamics,clark1995fluid} and targeted drug delivery in biomedical applications \cite{wu2024image,kara2020tumor,woodall2021patient}.
The Gaussian profile captures the physical intuition that the intensity of injection is highest at the source center and decays smoothly away from it in space.

\label{sec:model_problem}
\subsection{Finite Element Formulation of the Darcy-Transport System}
In this section, we will describe the numerical solution of the coupled PDE system. Here, we adopt a sequential strategy that takes advantage of the decoupled structure of the problem. Specifically, we first solve the steady-state Darcy flow system to compute the pressure and velocity fields, and then use the computed velocity as input to the time-dependent convection–diffusion equation governing solute transport. The finite element method (FEM) implementation is developed using the Firedrake framework \cite{rathgeber2016firedrake}. Since the source term $f(\mathbf{x})$ can become highly localized for small values of $ \sigma $, we employ an adaptive mesh refinement strategy to ensure sufficient spatial resolution near the source region.

We begin by formulating the steady-state Darcy system using a mixed finite element approach for the governing equations (\ref{eq:darcy})-(\ref{eq:div}) system. Let us introduce two distinct trial and test spaces for $\mathbf{v}$ and $p$. Let

$$
\begin{aligned}
\mathbf{V} &:= \left\{ \mathbf{w} \in H(\text{div}; \Omega) : \mathbf{w} \cdot \mathbf{n} = 0 \text{ on } \partial \Omega \right\}, \\
Q &:= L^2(\Omega),
\end{aligned}
$$

where $H(\text{div}; \Omega)$ denotes the space of square-integrable vector fields with square-integrable divergence. The homogeneous Neumann condition on the velocity field is imposed essentially via the definition of $\mathbf{V}$, while the Dirichlet condition on pressure is imposed naturally in the variational formulation.

The weak form then reads; find $(\mathbf{v}, p) \in \mathbf{V} \times Q$ such that:

$$
\begin{aligned}
(\mathbf{v}, \mathbf{w}) + \left(p, \nabla \cdot \mathbf{w} \right) &= 0 \quad &&\forall \mathbf{w} \in \mathbf{V}, \\
(\nabla \cdot \mathbf{v}, q) + (\alpha p, q) &= (\beta_1 f, q) \quad &&\forall q \in Q.
\end{aligned}
$$

To discretize the system, we select conforming finite-dimensional subspaces $\mathbf{V}_h \subset \mathbf{V}$ and $Q_h \subset Q$ such that the pair satisfies the Ladyzhenskaya–Babuška–Brezzi (LBB) condition. In our implementation, we use second-order Brezzi–Douglas–Marini (BDM2) elements for velocity, which provide $H(\text{div})$-conforming approximations with continuous normal components across element interfaces, and discontinuous piecewise linear (DG1) elements for pressure. This BDM2–DG1 pairing is a classical choice for mixed Darcy problems, known to ensure stability and optimal convergence \cite{brezzi1985two,boffi2013mixed}.

To generate \textit{reference solutions} to test the proposed model, we discretize the transport equation defined in Equation-\eqref{eq:transport} using a stabilized Galerkin finite element method \cite{quarteroni2009numerical, donea2003finite}. The Darcy velocity field $\mathbf{v}(x)$ is computed from the preceding system and treated as fixed input. We use the same adaptive mesh as that of the Darcy solver. Details of this mesh generation will be explained shortly. 

Regarding the transport equation \eqref{eq:transport}, let $\{\phi\} \subset V$ denote the set of admissible test functions for $V := H^1(\Omega)$. We 
consider a uniform temporal discretization with time step $\Delta t$, and approximate the solution $c(t,x)$ by $c^n(x) \approx c(t_n,x)$ at discrete tims $\{t_n\}$ using the backward Euler method. We seek $c^{n+1} \in V$ such that for all $\phi \in V$,
$$
\begin{aligned}
&\left( \frac{c^{n+1} - c^n}{\Delta t}, \phi \right) - (D \nabla c^{n+1}, \nabla \phi) + (\mathbf{v} c^{n+1}, \nabla \phi)
+ \tau \, (\mathbf{v} \cdot \nabla \phi, R^{n+1}(c)) = (\beta_2 f, \phi),
\end{aligned}
$$

where the streamline residual is defined as

$$
R^{n+1}(c) := \frac{c^{n+1} - c^n}{\Delta t} + \nabla \cdot (\mathbf{v} c^{n+1}) - \nabla \cdot (D \nabla c^{n+1}) - \beta_2 f,
$$

Stabilization parameter is computed elementwise as $\tau = \frac{h}{2 \|\mathbf{v}\|},$ where $h$ denotes the local cell diameter and $\|\mathbf{v}\|$ the magnitude of the advective velocity. The spatial discretization uses continuous piecewise linear (P1) finite elements over the shared adaptive mesh.

Lastly we will describe the adaptive mesh generation procedure. To accurately resolve the steep gradients induced by the localized source term $f(x)$, we generate an adaptively refined mesh that concentrates resolution near the centers of the Gaussian profiles. This is particularly important when the source width $\sigma$ is small, leading to highly localized source regions. The refinement procedure is implemented using Gmsh and designed to scale with the source configuration parameters \cite{geuzaine2009gmsh}.

Let $\{ \mathbf{x}_i \}_{i=1}^{N}$ denote the centers of the Gaussian source components, with corresponding spatial widths $\{ \sigma_i \}_{i=1}^{N}$. For each Gaussian, we define a target refinement radius $r_i$ and a transition width $\delta_i$ as:

$$
r_i = \max\left(k_i \sigma_i, \, 0.1 \cdot \min(L_x, L_y)\right), \qquad \delta_i = \frac{r_i}{2},
$$

where $k_i$ is a scaling factor that increases as $\sigma_i$ decreases, thus enforcing finer resolution around sharper sources. Specifically, $k_i$ is defined in the form of:

$$
k_i = k_0 + k_1 \cdot \frac{\sigma_{\text{ref}} - \sigma_i}{\sigma_{\text{ref}}},
$$

with $\sigma_{\text{ref}}$ a nominal reference scale and $k_0, k_1$ tunable constants. We set $\sigma_{\text{ref}}=1e-3, k_0=k_1=10$.

To smoothly transition between fine and coarse mesh regions, we define a background mesh size field $h(x, y)$ over the domain using a product of smooth decay functions centered at each source. For each source, we define:

$$
\psi_i(x, y) := \frac{1}{2} \left(1 + \tanh\left( \frac{ \| (x, y) - \mathbf{x}_i \| - r_i }{ \delta_i } \right) \right),
$$

and assemble the full mesh size field as:

$$
h(x, y) = h_{\min} + (h_{\max} - h_{\min}) \cdot \prod_{i=1}^N \psi_i(x, y),
$$

where $h_{\min}$ is a user-defined minimum mesh size and $h_{\max}$ is the maximum allowable element size. Our aim with this construction is to have $h(x, y) \approx h_{\min}$ near each Gaussian center and $h(x, y) \approx h_{\max}$ far from all sources while having smooth variation in between, avoiding abrupt transitions. In our implementation, we set  $h_{\min}=\min_{i} \frac{\sigma_i }{6}$ and $h_{max}=0.04\max(L_x,L_y)$.

\begin{figure}[H]
    \centering

    \begin{subfigure}{0.85\textwidth}
      \stackinset{l}{1mm}{t}{1mm}{\Large \color{white} \textbf{(a)}}{\fbox{\includegraphics[width=\linewidth]{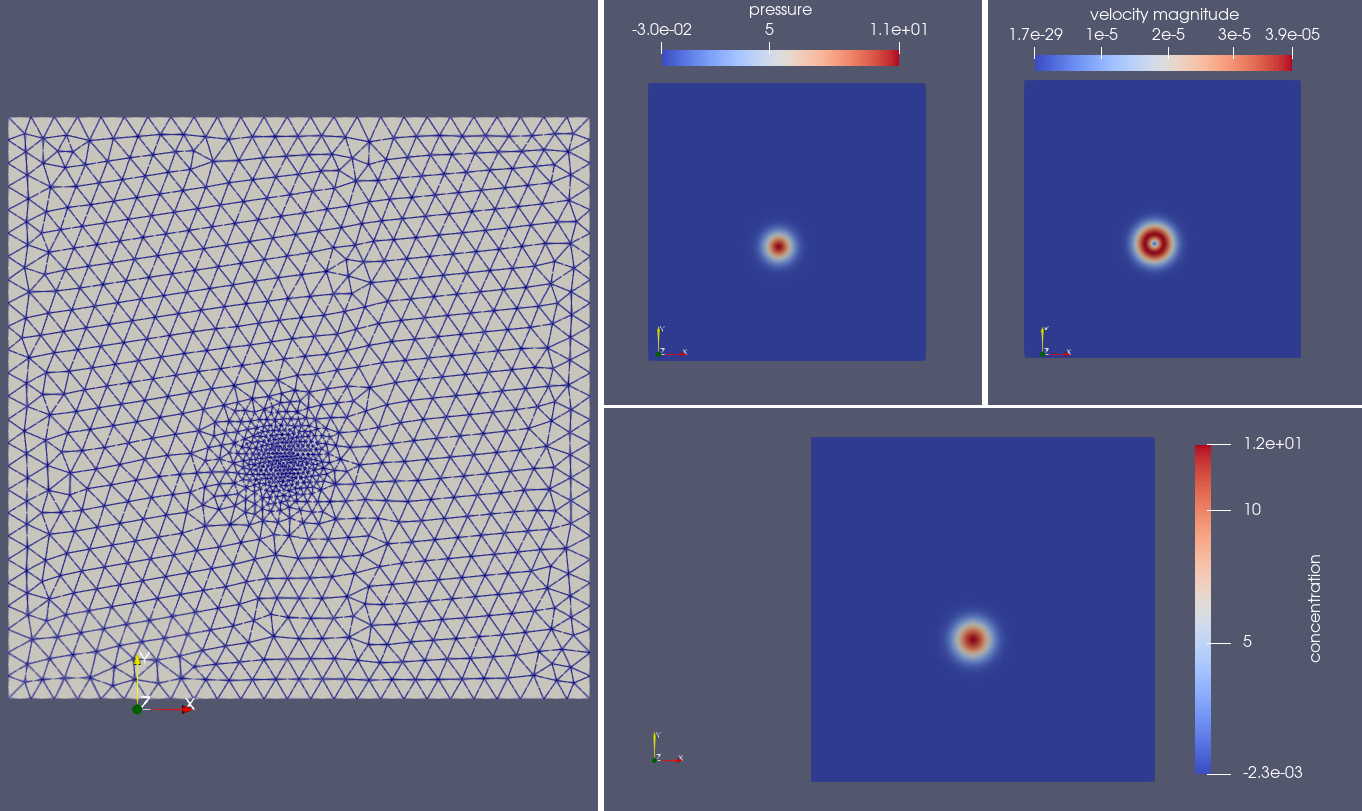}}}
    \end{subfigure} \vspace{4pt}
    
    \begin{subfigure}{0.85\textwidth}
      \stackinset{l}{1mm}{t}{1mm}{\Large  \color{white} \textbf{(b)}}{\fbox{\includegraphics[width=\linewidth]{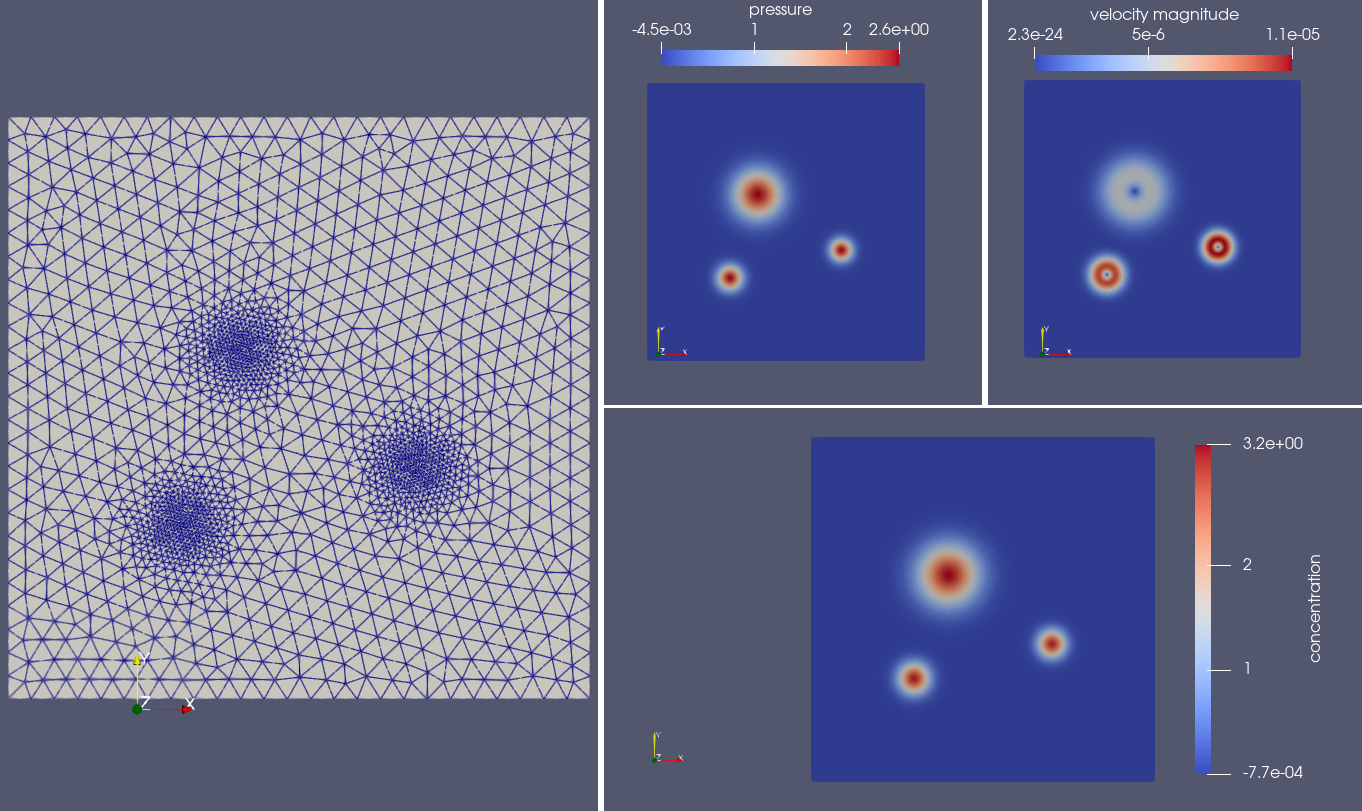}}}
    \end{subfigure}

\caption{
Representative results from the FEM pipeline for the coupled Darcy–Transport system at final time $T = 500$. Each row corresponds to a different Gaussian source configuration. The left panel shows the adaptive mesh generated based on the source centers. The right panel contains three scalar fields: the top-left shows the pressure field, the top-right shows the velocity magnitude, and the bottom shows the solute concentration profile.
\textbf{(a)} Single source with center $x_0 = (4.4,\, 3.73)$ and width $\sigma = 0.32$.
\textbf{(b)} Triple source with centers $x_0 = \{(4,6),\ (7,4),\ (3,3)\}$ and widths $\sigma = (0.59,\ 0.27,\ 0.30)$, respectively.
}

\label{fig:fem_flow}
\end{figure}

Figure~\ref{fig:fem_flow} illustrates the complete FEM pipeline, from adaptive mesh generation to the solution of the coupled Darcy–Transport system \eqref{eq:darcy}, \eqref{eq:div}, \eqref{eq:transport}, for representative single- and triple-source cases.

\label{sec:fem}

\section{Methodologies}
\label{sec:proposed_fram}

\subsection{Proposed Method}
\label{sec:proposed_method}

In this section, we outline our FEM–DeepONet coupling strategy and explain the mathematical and computational motivations behind its design. In principle, one could formulate the learning task as the operator mapping $f \mapsto (\mathbf{v}, p, c)$, seeking to predict the full velocity, pressure, and concentration fields from the source term. However, this approach presents both mathematical and computational challenges. First, the velocity and pressure fields are governed by the steady-state Darcy system, which is elliptic and typically well-conditioned. These equations can be solved efficiently and with high accuracy using a suitable numerical method(e.g. finite element method), as they converge rapidly and incur relatively low computational cost compared to the time-dependent transport problem. Consequently, attempting to learn the mapping $f \mapsto (\mathbf{v}, p)$ offers limited computational benefit.

More critically, the system exhibits a natural decoupling: the velocity field $\mathbf{v}$ enters the transport equation as an advective term, meaning that even small perturbations in the predicted $\mathbf{v}$ or $p$ fields can propagate nonlinearly into the concentration $c$, significantly amplifying errors. Learning the full coupled map $f \mapsto (\mathbf{v}, p, c)$ therefore becomes a considerably harder problem, both in terms of expressivity and stability. To address these issues, we propose a hybrid framework: we first solve the steady-state Darcy system using FEM to obtain accurate estimates of $\mathbf{v}$ and $p$, and then use this computed velocity field to drive the time-dependent convection–diffusion equation. We train DeepONet to learn the reduced mapping $f \mapsto c$, where the input incorporates both the source term $f$ and the advective field $\mathbf{v}$. This approach is particularly effective because the time-dependent transport equation constitutes the main computational bottleneck, while the steady Darcy system is comparatively cheap to solve numerically. 

To our knowledge, such a hybrid coupling of FEM solvers with operator-learning models like DeepONet has received little attention in the current literature. Beyond the clear computational advantages, this coupling offers several important benefits. First, it provides a modular framework, where the Darcy system can be solved with a high-fidelity, physics-based solver and seamlessly combined with a data-driven model for the more demanding transport component. This modularity allows the method to generalize more easily across heterogeneous media or varying flow conditions since these will be handled by FEM solver. Second, by explicitly solving the steady-state flow problem, the approach retains physical interpretability and numerical error control, avoiding the black-box nature of end-to-end learning and ensuring that discretization errors in the flow field can be systematically monitored. Third, the learning task becomes significantly more data-efficient, as the network no longer needs to implicitly learn the physics of fluid flow but can focus its capacity on approximating the transport dynamics. Finally,  framework offers transferability across applications: by adapting only the flow solver component, the overall system can be applied to a wide range of scientific and engineering problems.

The second key aspect of our study concerns the choice of spatial sampling strategy used during training, particularly on the trunk network side of the DeepONet architecture. In much of the recent operator learning literature, the training inputs are drawn from smooth function spaces, often using Gaussian random field (GRF) priors, and the solution is learned over a set of spatial points that are typically randomly sampled from a uniform distribution. This approach has been successful in many applications because the input functions are smooth and the solution varies relatively smoothly across the domain, allowing standard sampling from a uniform distribution to capture the main features of the solution.

However, in our setting, the source term $f(\mathbf{x})$ is modeled as a sharply localized Gaussian, representing a focused infusion into a porous medium. This leads to solution fields, particularly the concentration $c(\mathbf{x}, t)$, that exhibit steep gradients near the source, with fine-scale features that are poorly captured by uniform trunk sampling. As we will demonstrate in later sections, the classical random uniform sampling approach fails to resolve these sharp features. To address this challenge, we propose a simple yet effective adaptive trunk sampling strategy, which places a denser set of evaluation points near the regions of high variation induced by the Gaussian source. This ensures that the DeepONet accurately captures the local solution behavior where it is most sensitive, while avoiding unnecessary oversampling in smooth regions. In addition, we emphasize that unlike many prior studies, which rely on discrete input samples interpolated onto a fixed branch sensor grid, our Gaussian source functions are known analytically and can be evaluated directly on any desired grid. This eliminates the need for a separate mapping step on the branch side and allows us to focus our adaptive sampling efforts entirely on the trunk network.

\subsection{FEM-PI-DeepOnet Coupling with Adaptive Sampling}
\label{sec:coupling}
We now describe the complete workflow that integrates finite element computations with physics-informed DeepONet to construct the operator mapping

$$
f \mapsto c,
$$

where $f$ is the source function and $c$ is the resulting solute concentration field. As outline above, The steady-state Darcy system is solved using FEM to obtain the velocity field $v$, while the time-dependent convection–diffusion equation is learned by a PI-DeepONet model.

Let $\mathcal{A}$ and $\mathcal{B}$ be Banach spaces of input and output functions, respectively, and let

$$
\mathcal{G} : \mathcal{A} \to \mathcal{B}, \quad c(\mathbf{x},t) = \mathcal{G}(f)(\mathbf{x},t), \quad (\mathbf{x},t) \in \Omega \times[0,T],
$$

denote the operator mapping of interest. DeepONet approximates $\mathcal{G}$ using a neural network of the form:

$$
\mathcal{G}_\theta(f)(\mathbf{x},t) = \langle B_{\theta_b}(f(z_1), \dots, f(z_m)),\; T_{\theta_t}(\mathbf{x},t) \rangle,
$$

where:
\begin{itemize}
    \item $\{z_i\}_{i=1}^m \subset \Omega$ are fixed sensor points for evaluating the input function $f$,
    \item $B_{\theta_b} : \mathbb{R}^m \to \mathbb{R}^q$ is the branch network,
    \item $T_{\theta_t} : \Omega\times[0,T] \to \mathbb{R}^q$ is the trunk network,
    \item $\langle \cdot, \cdot \rangle$ denotes the Euclidean inner product in $\mathbb{R}^q$,
    \item $\theta = (\theta_b, \theta_t)$ denotes all trainable parameters.
\end{itemize}

In its original supervised form, DeepONet is trained using labeled pairs $\{(f^{(n)}, c^{(n)}(x))\}_{n=1}^N$. However, in the physics-informed setting, we exploit the differentiability of DeepONet outputs with respect to trunk inputs to impose PDE constraints directly, allowing unsupervised training without solution labels. This yields a Physics-Informed DeepONet (PI-DeepONet) that minimizes a composite residual loss over a collocation set in space–time \cite{wang2021learning}. We will describe this next.

To evaluate the PDE residuals during training, we construct a specialized set of trunk collocation points in space–time. The total collocation set is denoted by

$$
\mathcal{Y}_r = \mathcal{Y}_{\text{source}} \cup \mathcal{Y}_{\text{rand}} \cup \mathcal{Y}_{\text{bcs}} \cup \mathcal{Y}_{\text{ics}},
$$

where each subset serves a specific role.

\textbf{Source-Centered Residual Points}:  To resolve the steep gradients induced near the source, we place residual trunk points more densely in its vicinity using a polar sampling scheme:

$$
\mathcal{Y}_{\text{source}} = \left\{ (\mathbf{x}_i, t_i) \,\big|\, \mathbf{x}_i = \mathbf{x}_0 + r_j (\cos \theta_\ell, \sin \theta_\ell),\; r_j \in [0, 3\sigma],\; \theta_\ell \in [0, 2\pi),\; t_i \sim \mathcal{U}[0, T] \right\},
$$

This structured sampling yields $|\mathcal{Y}_{\mathrm{source}}| = n_r \cdot n_\theta$ where $n_r$ and $n_\theta$ denote the number of radial and angular subdivisions

\textbf{Random Residual Points}: To ensure broader spatial-temporal coverage, we include a uniformly random set over the domain:

$$
\mathcal{Y}_{\mathrm{rand}} = \{ (\mathbf{x}_i, t_i) \mid \mathbf{x}_i \sim \mathcal{U}(\Omega),\; t_i \sim \mathcal{U}[0, T] \}, \quad |\mathcal{Y}_{\mathrm{rand}}| = n_{\mathrm{rand}}.
$$

\textbf{Boundary and Initial Condition Points:} In addition to interior residual points, we generate collocation points to enforce Neumann boundary and initial conditions:
$$
\mathcal{Y}_{\mathrm{bcs}} = \{ (\mathbf{x}_j, t_j) \mid \mathbf{x}_j \sim \mathcal{U}(\partial \Omega),\; t_j \sim \mathcal{U}[0, T] \}, \quad |\mathcal{Y}_{\mathrm{bcs}}| = P_{\mathrm{bcs}},
$$
$$
\mathcal{Y}_{\mathrm{ics}} = \{ (\mathbf{x}_j, 0) \mid \mathbf{x}_j \sim \mathcal{U}(\Omega) \}, \quad |\mathcal{Y}_{\mathrm{ics}}| = P_{\mathrm{ics}}.
$$

\begin{figure}[H]
    \centering
    \includegraphics[width=\textwidth]{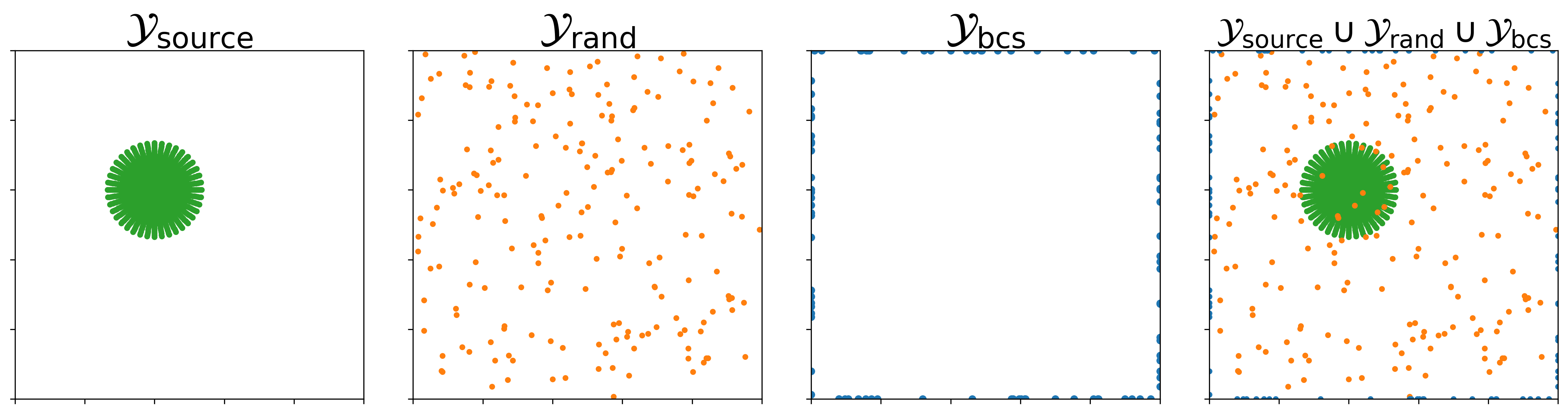}
    \caption{
        Visualization of trunk collocation point sampling for a single Gaussian source. 
        From left to right: (1) structured polar grid centered at the source, 
        $\mathcal{Y}_{\mathrm{source}}$; (2) uniformly sampled points over the domain, 
        $\mathcal{Y}_{\mathrm{rand}}$; (3) boundary condition points covering all four edges, 
        $\mathcal{Y}_{\mathrm{bcs}}$; and (4) the combined trunk set used during training.
    }
    \label{fig:trunk_grid}
\end{figure}

Figure-\ref{fig:trunk_grid} illustrates these three subsets for a representative Gaussian sampled.

The training process proceeds as follows. For each training instance:
\begin{enumerate}
    \item A Gaussian source function $f$ is sampled by drawing the center $\mathbf{x_0}$ and width $\sigma$.
    \item The steady-state Darcy system is solved using FEM on an adaptive mesh refined around $x_0$, yielding the velocity field $v(\mathbf{x})$.
    \item This velocity field and its derivatives are interpolated onto the union $\mathcal{Y}_{\text{source}} \cup \mathcal{Y}_{\text{rand}}$, and used in evaluating the PDE residuals.
    \item The source function is evaluated at a fixed set of branch sensor points $\{z_i\}_{i=1}^{m^2}$ to construct the input vector:

   $$
   \mathbf{b}^{(n)} = \left( f^{(n)}(z_1), \dots, f^{(n)}(z_{m^2}) \right) \in \mathbb{R}^{m^2}.
   $$
   \item  Each point $(\mathbf{x}_j, t_j) \in \mathcal{Y}_r^{(n)}$ is paired with $\mathbf{b}^{(n)}$ to evaluate the network output $c_\theta(\mathbf{x}_j, t_j)$ and its derivatives.
\end{enumerate}

Here, we used $c_\theta(\mathbf{x}, t) := \mathcal{G}_\theta(f)(\mathbf{x}, t)$ for simplicity. With this set-up, each training instance is thus a tuple $\left(\mathbf{b}^{(n)}, \mathcal{Y}_r^{(n)}\right)$, and the full training dataset can we written as:

$$
\mathcal{D}_{\text{train}} = \left\{ \left( \mathbf{b}^{(n)}, \mathcal{Y}_r^{(n)} \right) \right\}_{n=1}^N.
$$

PI-DeepONet is constructed in a similar manner to PINNs. In this sense,  a composite loss function can be defined as follow

\begin{align}
\mathcal{L}_{\text{total}}(\theta) &= \lambda_{\text{res}} \mathcal{L}_{\text{res}}(\theta) + \lambda_{\text{bcs}} \mathcal{L}_{\text{bcs}}(\theta) + \lambda_{\text{ics}} \mathcal{L}_{\text{ics}}(\theta),
\label{eq:total_loss}
\end{align}

with fixed weights $\lambda_{\text{res}}, \lambda_{\text{bcs}}, \lambda_{\text{ics}}$ chosen empirically.

The first term corresponds to the residual of the governing PDE:
$$
\mathcal{L}_{\text{res}}(\theta) = \frac{1}{|\mathcal{Y}_{\text{res}}|} \sum_{(\mathbf{x},t) \in \mathcal{Y}_{\text{res}}} \mathcal{R}_\theta(\mathbf{x}, t)^2,
$$
where
$$
\mathcal{R}_\theta(\mathbf{x}, t) = \frac{\partial c_\theta}{\partial t}(\mathbf{x}, t) - \nabla \cdot (D \nabla c_\theta(\mathbf{x}, t)) + \nabla \cdot (v(\mathbf{x})\, c_\theta(\mathbf{x}, t)) - \beta_2 f(\mathbf{x})
$$
is the residual associated with the convection–diffusion equation \eqref{eq:transport}.

The second term penalizes violations of the homogeneous Neumann boundary condition:
$$
\mathcal{L}_{\text{bcs}}(\theta) = \frac{1}{|\mathcal{Y}_{\text{bcs}}|} \sum_{(\mathbf{x},t) \in \mathcal{Y}_{\text{bcs}}} \left( \mathbf{n} \cdot \nabla c_\theta(\mathbf{x}, t) \right)^2,
$$
where $\mathbf{n}$ denotes the outward unit normal at the boundary point $\mathbf{x}$, computed analytically from the rectangular domain geometry (e.g., $\mathbf{n} = (1, 0)$ on the right edge, $\mathbf{n} = (0, -1)$ on the bottom, etc.).

The third and final component enforces the initial condition:
$$
\mathcal{L}_{\text{ics}}(\theta) = \frac{1}{|\mathcal{Y}_{\text{ics}}|} \sum_{\mathbf{x} \in \mathcal{Y}_{\text{ics}}} c_\theta(\mathbf{x}, 0)^2.
$$

We should note that, in \cite{wang2022improved}, authors incorporated a weighting strategy inspired by Neural Tangent Kernel (NTK) theory to balance the magnitude of gradients across residual, boundary, and initial loss terms. This adaptive scheme is shown to accelerate convergence and improve predictive accuracy, particularly in fully self-supervised settings in expense of longer training times. In our case, however, we impose Neumann boundary conditions on the concentration field, and we observed that the directional gradients required by NTK-based weighting become ill-defined or unstable near the domain edges. As a result, we do not adopt the NTK reweighting strategy. Instead, we use a fixed weighting scheme for the composite loss function as defined in Equation~\eqref{eq:total_loss}, with
$\lambda_{\mathrm{res}} = 10.0$, $\lambda_{\mathrm{bcs}} = 10^{-3}$, and $\lambda_{\mathrm{ics}} = 1.0$.

\begin{figure}[H]
    \centering
    \includegraphics[width=\textwidth]{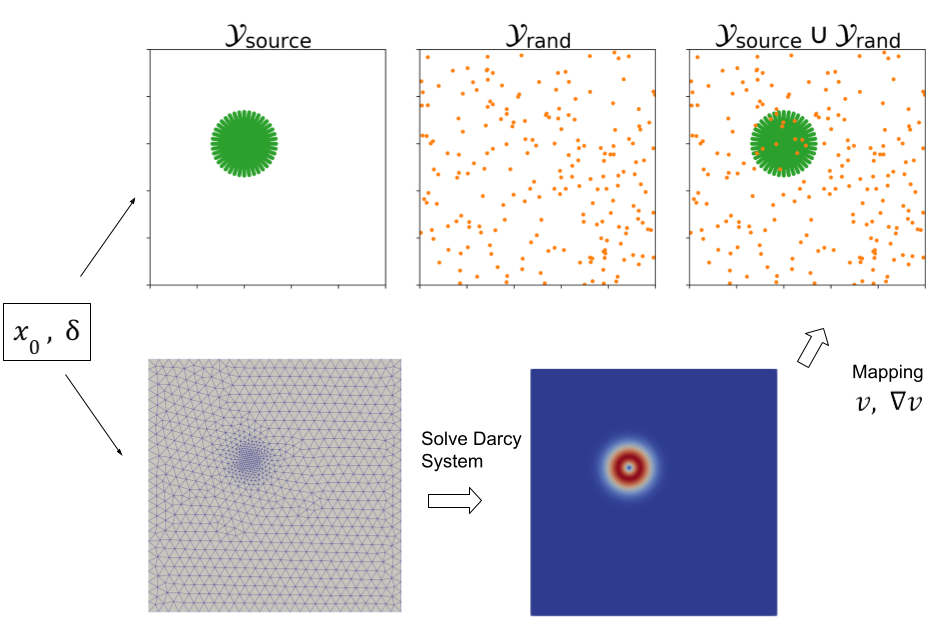}
    \caption{Overview of the FEM–DeepONet coupling strategy. Given a sampled Gaussian source term defined by parameters $\mathbf{x}_0$ and $\sigma$, an adaptive mesh is generated around the source region, and the Darcy system is solved via the finite element method (FEM). The resulting velocity field $\mathbf{v}(\mathbf{x})$ and its gradient $\nabla \mathbf{v}(\mathbf{x})$ are then interpolated onto the trunk collocation points $\mathcal{Y}{\mathrm{source}} \cup \mathcal{Y}{\mathrm{rand}}$, which are used to evaluate the PDE residuals in the PI-DeepONet loss. For simplicity, we display velocity magnitude here in the second image in bottom row.}
    \label{fig:flow}
\end{figure}

Fig-\ref{fig:flow} illustrates end-to-end training pipeline for the proposed method.

\begin{figure}[H]
    \centering
    \includegraphics[width=\textwidth]{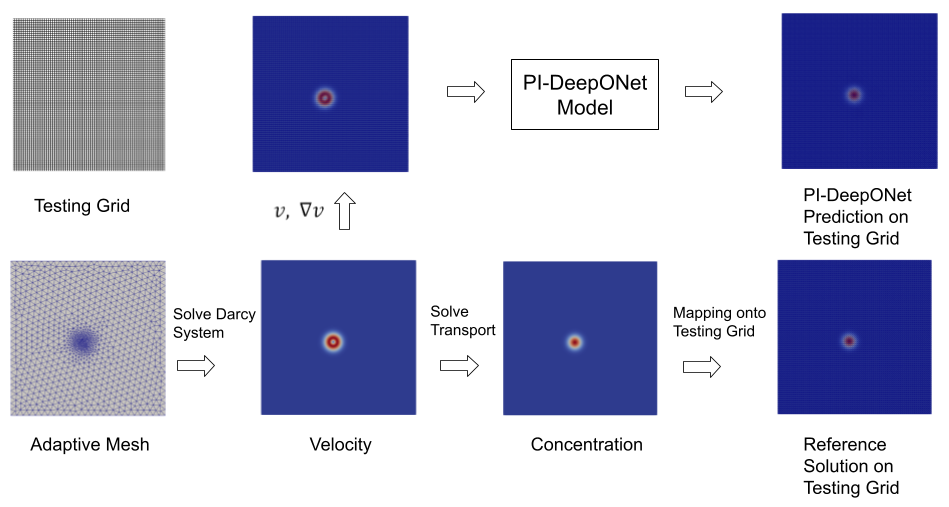}
\caption{Overview of the testing procedure for evaluating PI-DeepONet predictions against FEM reference solutions. Each row corresponds to a different pipeline applied to the same sampled Gaussian source. \textbf{Bottom row (FEM):} (1) An adaptive mesh is generated around the source location. (2) The steady-state Darcy system is solved to compute the velocity field. (3) This velocity is used to solve the time-dependent convection–diffusion equation on the same mesh. (4) The resulting concentration field is interpolated onto a fixed $k \times k$ Cartesian grid to obtain the reference solution. \textbf{Top row (DeepONet):} (1) The FEM-computed velocity field is interpolated onto the $k \times k$ grid. (2) This velocity, along with the source function, is passed through the PI-DeepONet model to predict the solute concentration. The resulting prediction is compared to the FEM reference to assess model performance. Note: The illustration simplifies the inputs to highlight flow; full model input structure is described in Section-\ref{sec:coupling}}
    \label{fig:testing_flow}
\end{figure}

\textbf{Testing Procedure:} To evaluate the model's predictive accuracy, we compare the PI-DeepONet outputs against finite element (FEM) solutions computed on an adaptive mesh. Both predictions and reference solutions are mapped onto a uniform $k \times k$ Cartesian grid to ensure consistency. For each test source, the FEM pipeline generates a high-fidelity concentration field by solving the Darcy and transport equations sequentially, followed by interpolation onto the evaluation grid. In parallel, the PI-DeepONet model uses the same source and FEM-computed velocity field to generate its prediction directly on the same grid. We illustrate the full testing pipeline in Figure-\ref{fig:testing_flow}.

\subsection{Model Architecture and Baseline Parameters}
In this work, we adopt the modified DeepONet architecture introduced in \cite{wang2022improved}, which we discussed in the introduction. Unless otherwise stated, both the branch and trunk networks consist of 4 hidden layers with 128 neurons per layer. All weights are initialized using the Glorot normal scheme, and hyperbolic tangent (tanh) activations are used. Training is conducted using mini-batch stochastic gradient descent via the Adam optimizer, with exponential learning rate decay. We set the initial learning rate to $10^{-3}$, decay steps to 5000, and decay rate to 0.95. All models are trained for $3\times 10^5$ iterations with a batch size of 200. While preliminary tests showed that fewer epochs (e.g., 300k–600k) were sufficient for simpler cases, we adopted a uniform upper-bound setting to ensure convergence across all configurations and to simplify comparison.

Table~\ref{tab:pde_params} summarizes the physical parameters used in the governing PDE system, while Table~\ref{tab:deeponet_params} lists the key modeling and architectural settings used in the PI-DeepONet experiments.

\begin{table}[H]
\centering
\begin{tabular}{|l|l|l|p{7cm}|}
\hline
\textbf{Parameter} & \textbf{Value} & \textbf{Unit} & \textbf{Explanation} \\
\hline \hline
% % Problem Configuration
$\Omega$ & $[0,10] \times [0,10]$ & mm & Spatial domain \\
K & $10^{-9}$ & mm$^2$ & Intrinsic permeability \\
$\mu$ & $9 \times 10^{-4}$ & Pa $\cdot$ s & Viscosity of water \\
$\alpha$ & $10^{-2}$ & - & Sink term \\
D & $4 \times 10^{-6}$ & mm$^2$/s & Diffusion coefficient \\
$\beta_1$ & $\frac{5}{60}$ & mm$^2$/s & Fluid addition rate \\
$\beta_2$ & $\frac{5}{240}$ & mmol/s & Concentration addition rate \\
\hline
\end{tabular}
\caption{Physical parameters used in the Darcy and convection–diffusion PDE system.}
\label{tab:pde_params}
\end{table}

We should note that to keep the training pipeline lightweight and modular, we use set of fixed sampling-related hyperparameters across all experiments. The values provided in Table-\ref{tab:deeponet_params} were found to provide a good trade-off between accuracy and computational cost in both single and multi-source settings, which will be detailed in the subsequent section.

\begin{table}[H]
\centering
\begin{tabular}{|l|l|p{10cm}|}
\hline
\textbf{Parameter} & \textbf{Value} & \textbf{Explanation} \\
\hline \hline
$N$ & 2000 & Number of Gaussian input functions sampled \\
$m$ & $30 \times 30$ & Number of branch sensor points (grid) \\
$P_{\mathrm{bcs}}$ & 100 & Number of boundary condition (BC) collocation points \\
$P_{\mathrm{ics}}$ & 5 & Number of initial condition (IC) collocation points \\
$n_r$ & 30 & Radial points around the source ($r \leq 3\sigma$) \\
$n_\theta$ & 30 & Angular points around the source \\
$n_{\mathrm{rand}}$ & 300 & Number of random collocation points \\
$N_{test}$ & 30 & Number of input functions used for model testing \\
$k$ & $80 \times 80$ & dimensions of the testing grid \\
\hline
\end{tabular}
\caption{Sampling configurations used in the PI-DeepONet training pipeline.}
\label{tab:deeponet_params}
\end{table}

\textbf{Error Metrics}: All evaluation metrics are computed on a uniform $80 \times 80$ Cartesian spatial grid over the domain $\Omega = [0,10]\times [0,10]$, and at $N_t=10$ time steps in the interval $(0, T]$. The test set consists of $N_{\text{test}} = 30$ independently sampled source functions.

Let $c(\mathbf{x}, t)$ denote the ground truth concentration field obtained from the FEM solver, and $c_\theta(\mathbf{x}, t)$ denote the prediction from the trained PI-DeepONet model with parameters $\theta$, as defined earlier. We define two relative error metrics computed over all test instances of interest:

\begin{subequations}
\begin{align}
\mathcal{E}_{\text{full}} &= \frac{1}{N_{\text{test}}} \sum_{i=1}^{N_{\text{test}}} \frac{ \| c^{(i)}_\theta - c^{(i)} \|_2 }{ \| c^{(i)} \|_2 }, \\
\mathcal{E}_{T} &= \frac{1}{N_{\text{test}}} \sum_{i=1}^{N_{\text{test}}} \frac{ \| c^{(i)}_\theta(\cdot, T) - c^{(i)} (\cdot, T) \|_2 }{ \| c^{(i)}(\cdot, T) \|_2 },
\end{align}
\end{subequations}

where $\| \cdot \|_2$ denotes the standard Euclidean norm over the relevant grid. The metric $\mathcal{E}_{\text{full}}$ captures the model’s overall fidelity in approximating the full spatiotemporal evolution of the solution. Since our primary interest lies in accurately capturing the final concentration profile at $t = T$, we also report $\mathcal{E}_{T}$, which quantifies the model’s accuracy at the final time step. A detailed discussion of this design choice appears in Section-\ref{sec:discussion}.

\section{Results}

In this section, we consider several configurations for evaluating the operator learning capabilities of the proposed PI-DeepONet pipeline. Our primary focus is the mapping from a single Gaussian source to the concentration field at the final time, with the center and width of the Gaussian are randomly sampled. Building on this foundation, we extend the setup to more complex source terms defined as mixtures of multiple Gaussians. These multi-source functions are expressed as normalized sums of individual Gaussians, each with its own center and width:

$$
f(\mathbf{x}) = \frac{1}{\mathcal{N}} \sum_{i=1}^{N_{\text{source}}} \exp\left( -\frac{ \lVert \mathbf{x} - \mathbf{x}_i \rVert^2 }{2\sigma_i^2} \right),
$$

where $ \mathbf{x}_i \sim \mathcal{U}(\Omega_{\text{src}}) $ and $ \sigma_i \sim \mathcal{U}(\sigma_{\min}, \sigma_{\max}) $ denote the center and spatial width of the $ i $-th Gaussian, drawn independently from uniform distributions. Here, $ \Omega_{\text{src}} \subset \Omega $ specifies a source-specific region within the domain, and $ \mathcal{N} $ is a normalization constant. However, jointly varying both the center and the width across multiple sources significantly increases the complexity of the learning task. Specifically, the number of possible source configurations grows exponentially with the number of Gaussians, and the resulting input functions become increasingly difficult to represent due to overlapping supports, varying spatial scales, and nontrivial interaction patterns.

To explore what can be achieved under the same sampling and training constraints used in the single-source case, we focus on three practical subcases where a constraint is introduced into the source configuration. In the first, each training instance contains either one or two Gaussians, randomly selected, allowing the model to experience a range of source complexities. In the second, we consider the case of exactly three fixed-width Gaussians per sample. In the third subcase, we fix the positions of the three Gaussians and instead allow their widths to vary.
From a practical standpoint, fixing the Gaussian width corresponds to using a consistent injection mechanism or fluid dispersal device with known spatial characteristics while fixing the centers reflects repeatable positioning of hardware or predefined injection zones.

\subsection{Single Source with Varying Center and Width}
\label{sec:single_source}
% \erdi{
% \begin{itemize}
%     \item Report the results for this case with $T=50,250,500$.
%     \item Conduct a simple ablation to show smart sampling helps.
%     \item How fast is this procedure compared to FEM?
% \end{itemize}
% }

We begin by evaluating the model’s performance in the baseline setting, where the source term consists of a single Gaussian profile with randomized spatial center and width. The source function is given by:

$$
f(\mathbf{x}) = \frac{1}{\mathcal{N}} \exp\left( -\frac{ \lVert \mathbf{x} - \mathbf{x}_0 \rVert^2 }{2\sigma^2} \right),
$$

where the center $\mathbf{x}_0 \sim \mathcal{U}(\Omega_{\text{src}})$ is drawn uniformly from a square subregion $\Omega_{\text{src}} = [3,7] \times [3,7] \subset \Omega$, and the spatial width $\sigma \sim \mathcal{U}(0.25, 0.60)$. This means its effective support (within $3 \sigma$) \textit{covers on average only about 3.7\% of the total domain area}.

\begin{figure}[H]
    \centering
    \begin{subfigure}{0.45\textwidth}
      \stackinset{l}{0.5mm}{t}{0.5mm}{\large \color{black} \textbf{(a)}}{\fbox{\includegraphics[width=\linewidth]{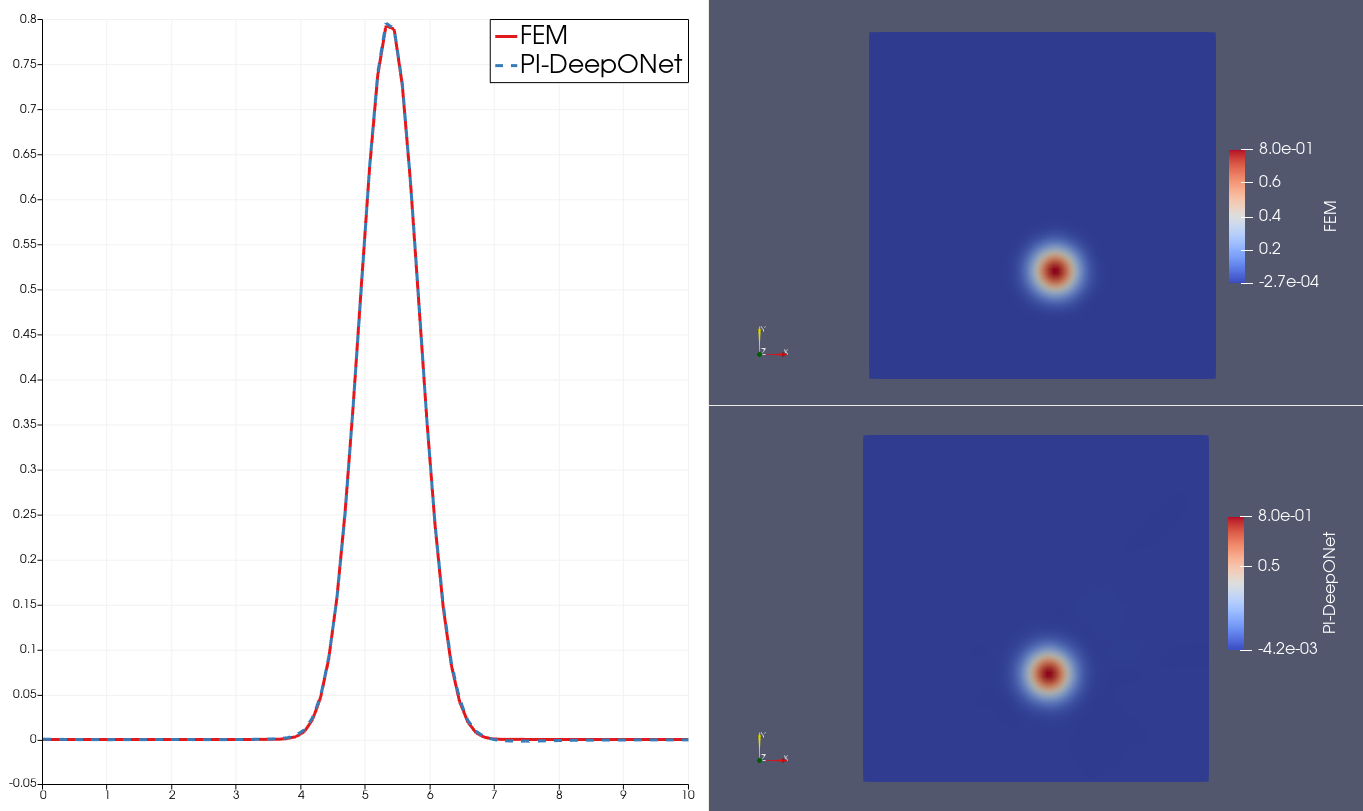}}}
    \end{subfigure} \hspace{4pt}
    \begin{subfigure}{0.45\textwidth}
      \stackinset{l}{1mm}{t}{1mm}{\large  \color{black} \textbf{(b)}}{\fbox{\includegraphics[width=\linewidth]{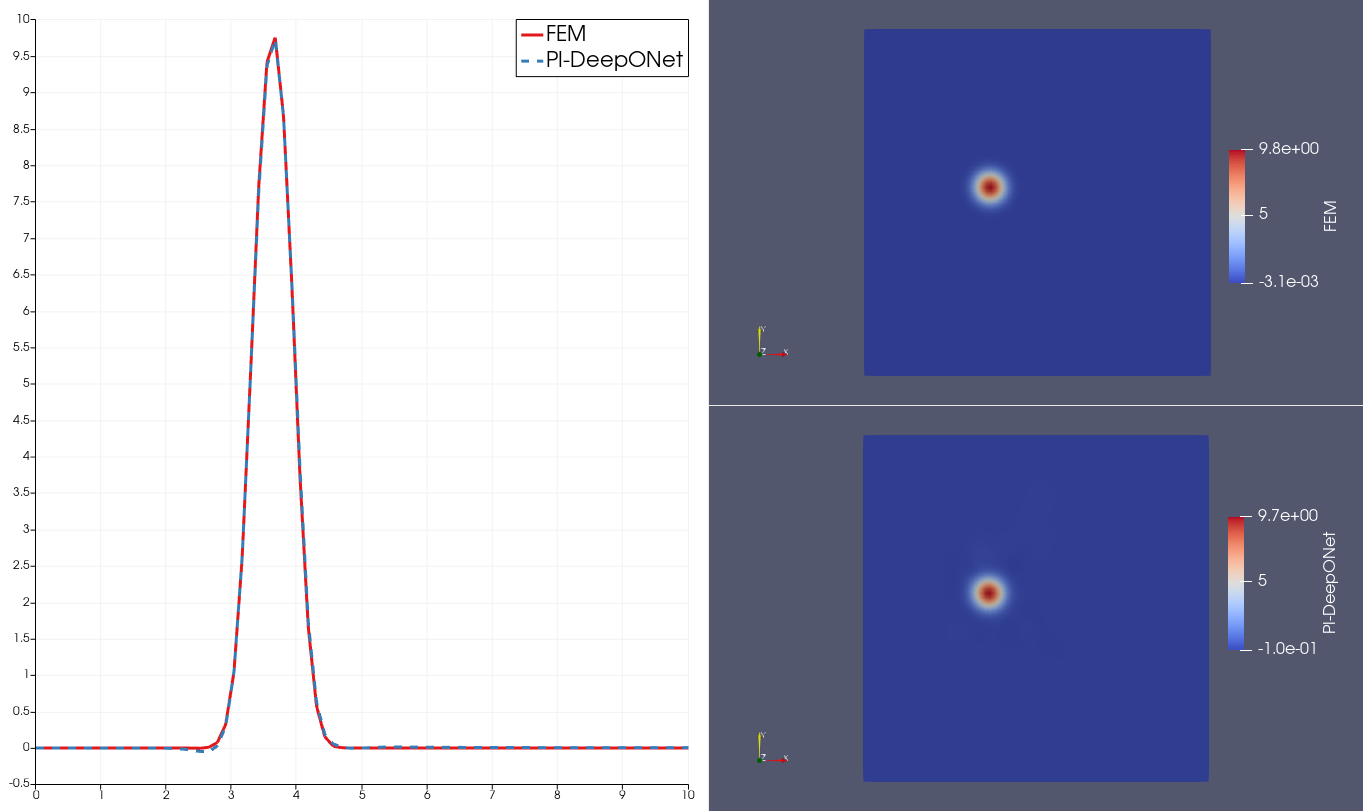}}}
    \end{subfigure}
    
    \vspace{5pt}
    
    \begin{subfigure}{0.45\textwidth}
      \stackinset{l}{1mm}{t}{1mm}{\large  \color{black} \textbf{(c)}}{\fbox{\includegraphics[width=\linewidth]{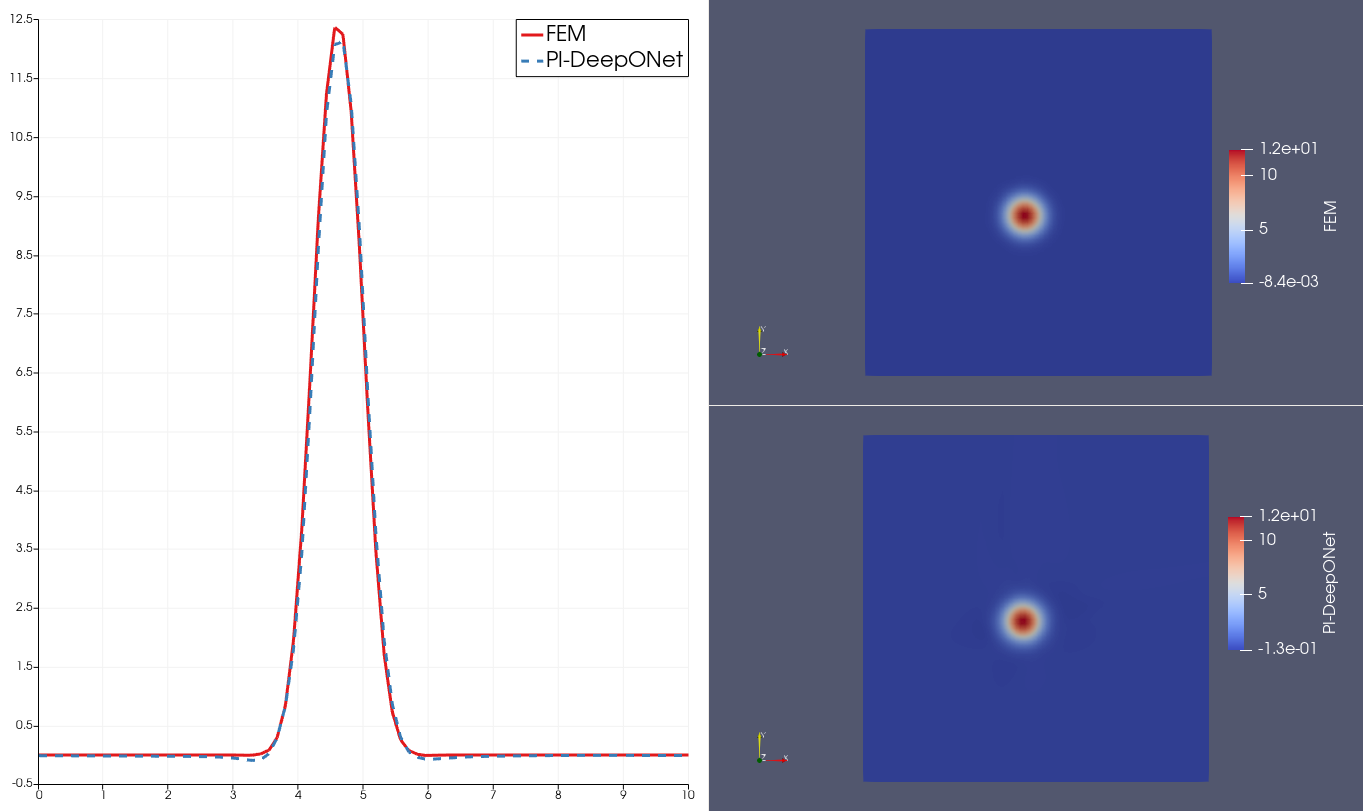}}}
    \end{subfigure}
\caption{
Comparison of PI-DeepONet and FEM solutions for three representative test cases at final times $T = 50$, $250$, and $500$. Each panel shows: 1D slice passing through the $y$-center of the source at $t = T$.  
Right: the full spatial concentration field from FEM (top) and PI-DeepONet (bottom), both at time $t = T$.
\textbf{(a)} $T = 50, x_0 = (5.37,\, 3.11), \sigma = 0.45$
\textbf{(b)} $T = 250$: source centered at $x_0 = (3.64,\, 5.44)$ with width $\sigma = 0.27$
\textbf{(c)} $T = 500, x_0 = (4.61,\, 4.62), \sigma = 0.34$.
}
\label{fig:single_gauss}
\end{figure}

PI-DeepONet is able to recover the final-time concentration field with high accuracy. As shown in Table-\ref{tab:single_source_errors}, the mean relative error across all time steps remains around $10\%$, while the final-time error drops is around 4\% for moderate and long time horizons. Figure-\ref{fig:single_gauss} illustrates representative comparisons between the predicted and FEM-simulated concentration fields for several test cases, showing close agreement in both shape and peak location. 

\begin{table}[H]
\centering
\begin{tabular}{|l|l|l|}
\hline
\textbf{Final Time} & $\mathcal{E}_{\text{full}}$ & $\mathcal{E}_{T}$ \\
\hline
$T=50$  & 9.99\%  & 1.97\% \\ \hline
$T=250$ & 10.06\% & 2.72\% \\ \hline
$T=500$ & 10.08\% & 2.86\% \\ \hline
\end{tabular}
\caption{Relative L$^2$ errors for single-source test cases. Here, $\mathcal{E}_{\text{full}}$ denotes the average error across all time steps, and $\mathcal{E}_T$ denotes the error at final time $t = T$.}
\label{tab:single_source_errors}
\end{table}

\subsection{Random Sampling of One or Two Sources with Fixed Width}
\label{sec:double_source}
We next consider a controlled extension of the baseline setting by allowing each input function to consist of either one or two Gaussian sources, randomly selected for each training and test instance. In this setup, the spatial width of each Gaussian is held fixed at a representative value $\sigma = 0.45$, while the centers are drawn independently from the same region as before: $\mathbf{x}_i \sim \mathcal{U}(\Omega_{\text{src}})$, with $\Omega_{\text{src}} = [3,7] \times [3,7]$. The source function is defined as:

$$
f(\mathbf{x}) = \frac{1}{\mathcal{N}} \sum_{i=1}^{N_{\text{source}}} \exp\left( -\frac{ \lVert \mathbf{x} - \mathbf{x}_i \rVert^2 }{2\sigma^2} \right),
$$

where $N_{\text{source}} \in \{1, 2\}$ is sampled uniformly, and $\mathcal{N}$ denotes a normalization factor ensuring consistent amplitude across inputs. 

As summarized in Table-\ref{tab:double_source_errors}, the model performs remarkably well in this setting, achieving final-time prediction accuracy comparable to the single-source case. The PI-DeepONet learns to generalize across a diverse set of overlapping and spatially separated source profiles without requiring a larger model or increased collocation budget. This formulation is also preferable to training a separate model with exactly two fixed sources, as it introduces less structural rigidity and better reflects real-world variability. In practical terms, it corresponds to scenarios where the number of injection sites may vary between deployments or trials, e.g., due to experimental error, adaptive control, or operational decisions

\begin{figure}[H]
    \centering

    \begin{subfigure}{0.45\textwidth}
      \stackinset{l}{0.5mm}{t}{0.5mm}{\large \color{black} \textbf{(a1)}}{\fbox{\includegraphics[width=\linewidth]{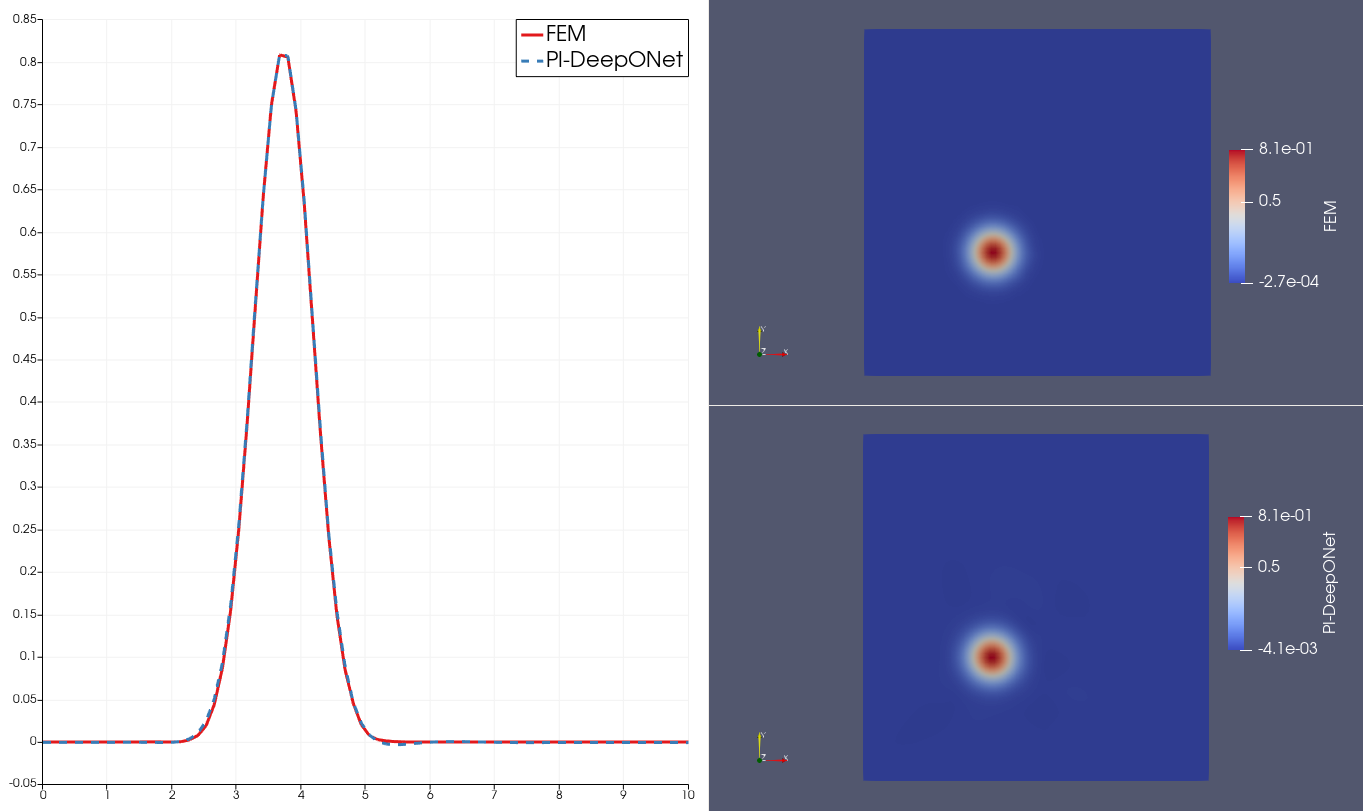}}}
    \end{subfigure} \hspace{4pt}
    \begin{subfigure}{0.45\textwidth}
      \stackinset{l}{1mm}{t}{1mm}{\large  \color{black} \textbf{(a2)}}{\fbox{\includegraphics[width=\linewidth]{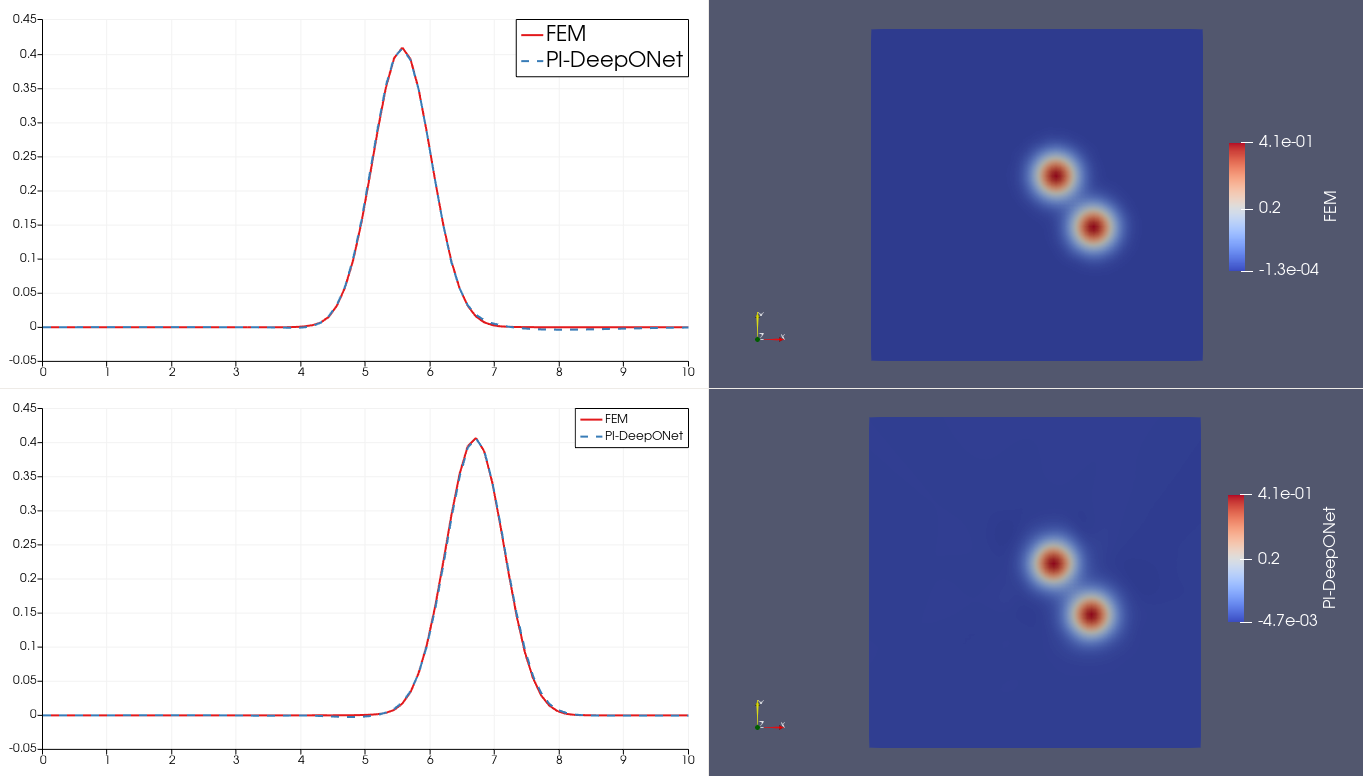}}}
    \end{subfigure}
    
    \vspace{5pt}
    
    \begin{subfigure}{0.45\textwidth}
      \stackinset{l}{1mm}{t}{1mm}{\large  \color{black} \textbf{(b1)}}{\fbox{\includegraphics[width=\linewidth]{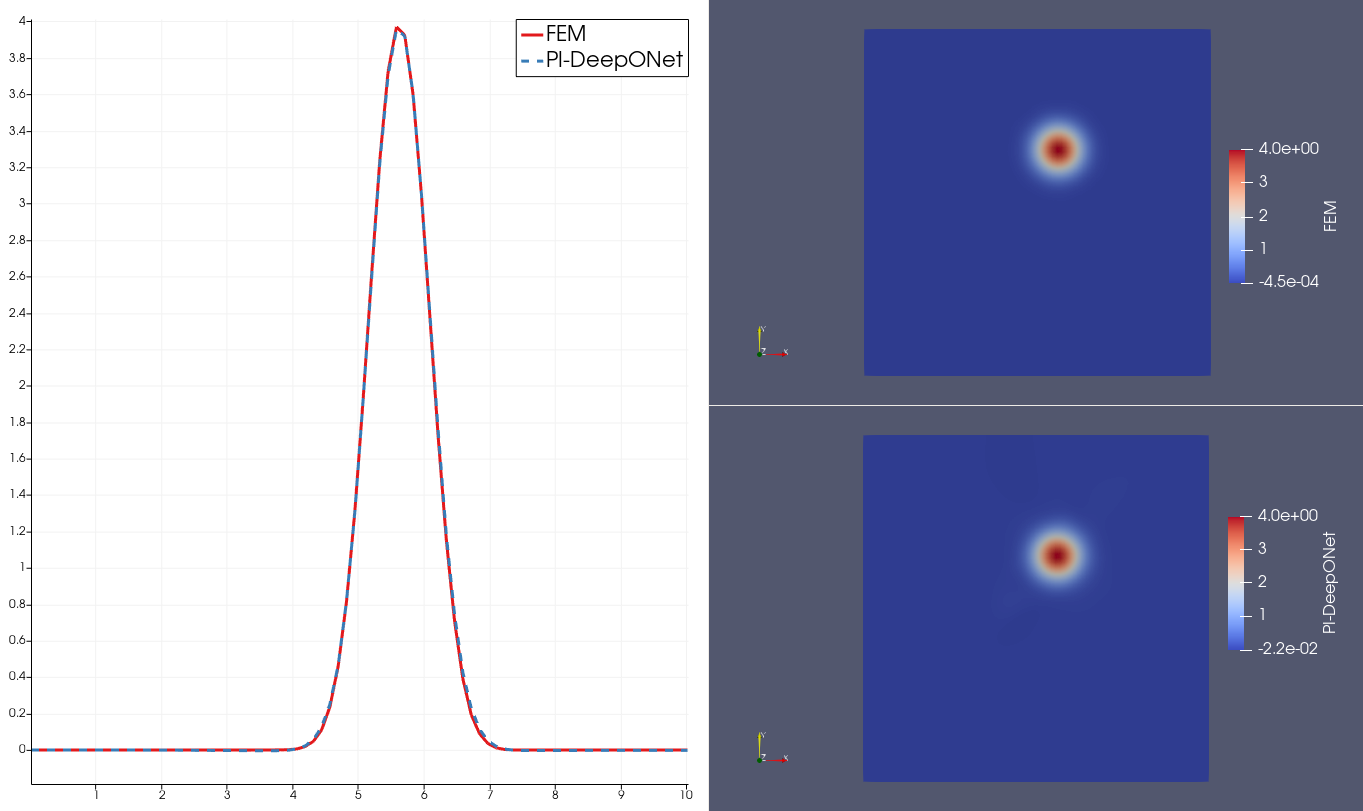}}}
    \end{subfigure}\hspace{4pt}
        \begin{subfigure}{0.45\textwidth}
      \stackinset{l}{1mm}{t}{1mm}{\large  \color{black} \textbf{(b2)}}{\fbox{\includegraphics[width=\linewidth]{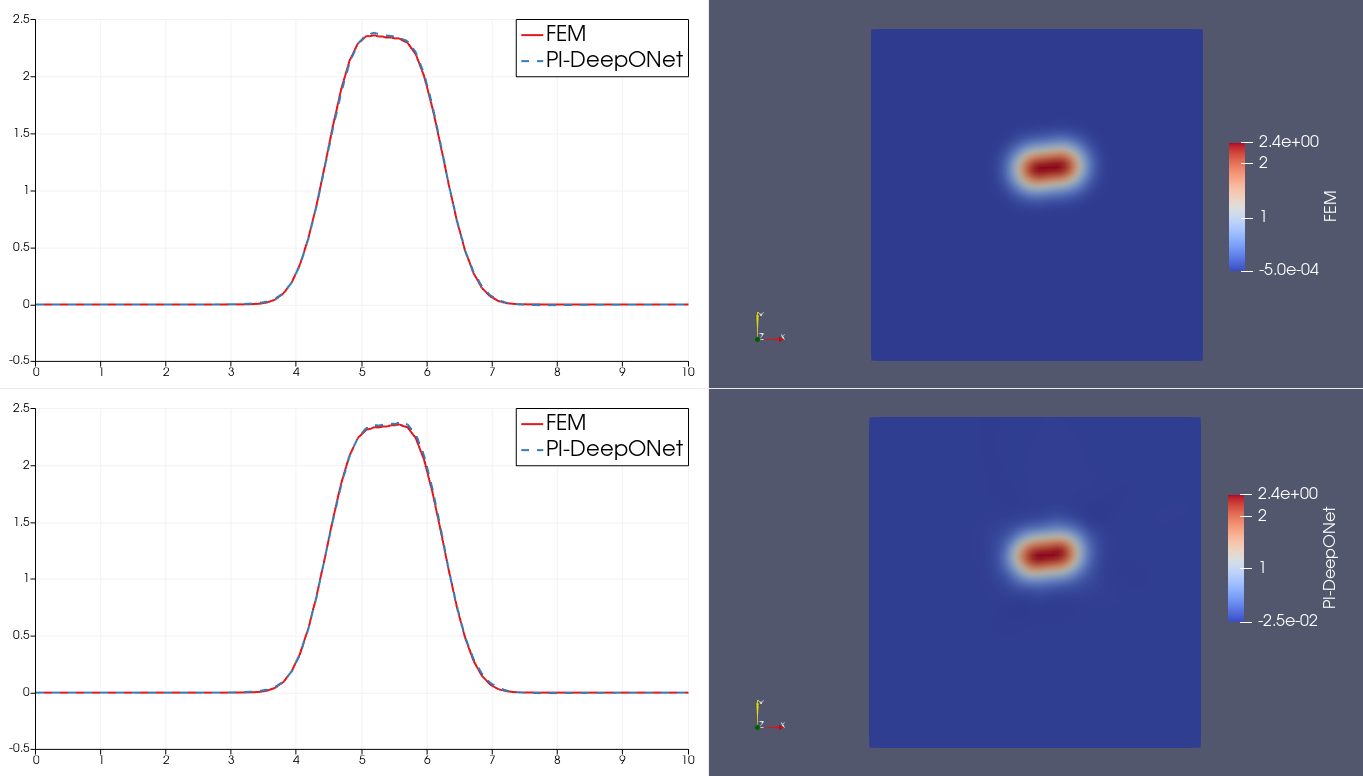}}}
    \end{subfigure}
    
\vspace{5pt}
        \begin{subfigure}{0.45\textwidth}
      \stackinset{l}{1mm}{t}{1mm}{\large  \color{black} \textbf{(c1)}}{\fbox{\includegraphics[width=\linewidth]{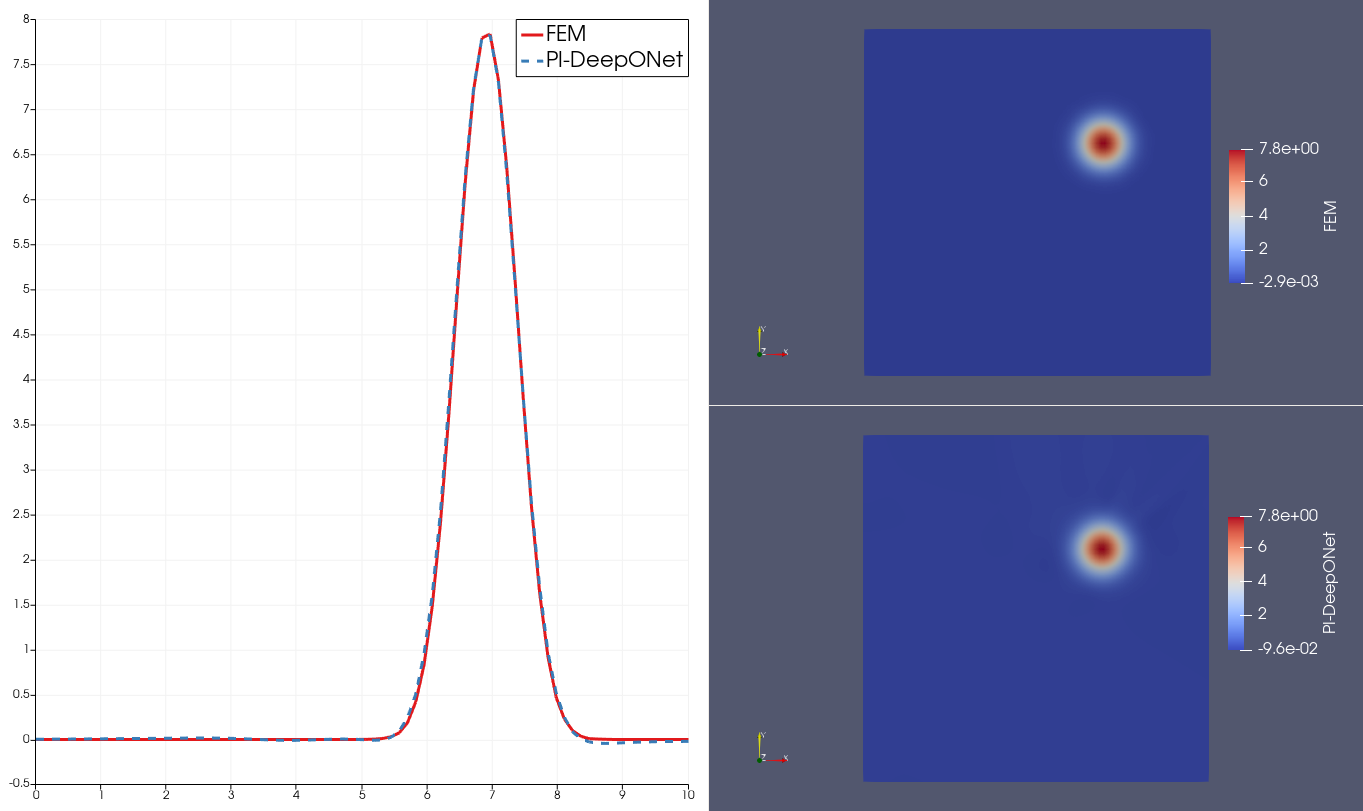}}}
    \end{subfigure}\hspace{4pt}
        \begin{subfigure}{0.45\textwidth}
      \stackinset{l}{1mm}{t}{1mm}{\large  \color{black} \textbf{(c2)}}{\fbox{\includegraphics[width=\linewidth]{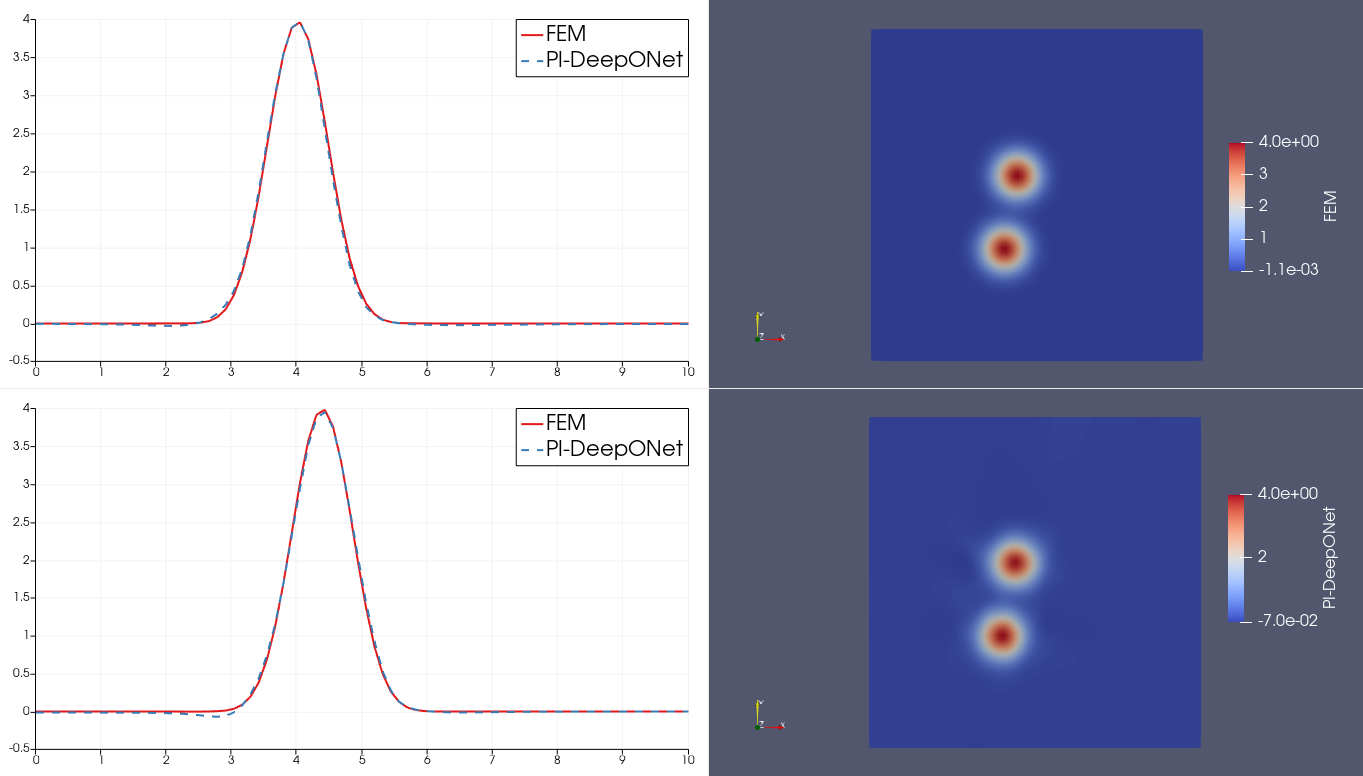}}}
    \end{subfigure}
\caption{
Comparison of PI-DeepONet and FEM solutions for three test cases at final times $T = 50$, $250$, and $500$, with each row corresponding to one of the final times, respectively. In each row:  \textbf{Left Image}: a 1D slice passing through the $y$-center of a single Gaussian source.  
\textbf{Right Image}: two 1D slices passing through the $y$-centers of the \textit{first} and \textbf{second} Gaussian sources, respectively.  
In all panels, the full spatial concentration field from FEM (top) and PI-DeepONet (bottom) is shown on the right of each sub-image.
\textbf{First Row ($T = 50$)}  
\textbf{(a1)}: single source at $x_0 = (3.73,\,3.56)$,  
\textbf{(a2)}: double source with centers $x_0 = [(5.57,5.58),(6.69,4.02)]$.
\textbf{Second Row ($T = 250$)}  
\textbf{(b1)}: single source at $x_0 = (5.61,6.52)$,  
\textbf{(b2)}: double source with centers  $x_0 = [(4.89,5.77),(5.82,5.88)]$.
\textbf{Third Row ($T = 500$)}  
\textbf{(c1)}: first source at $x_0 = (6.91,6.71)$,  
\textbf{(c2)}: double source with centers  $x_0 = [(4.02,3.37),(4.40,5.58)]$.
}
\label{fig:random_double_source}
\end{figure}

Figure-\ref{fig:random_double_source} provides representative examples across final times $T=50,250,500$ showing model predictions for both one-source and two-source cases sampled at random. For example, in the second row (panel b2), the two Gaussians are located very close to one another and exhibit substantial overlap. Despite this configuration, the model still accurately reconstructs the final concentration field, capturing both the peak structure and spatial extent of the true solution.

\begin{table}[H]
\centering
\begin{tabular}{|l|l|l|}
\hline
\textbf{Final Time} & $\mathcal{E}_{\text{full}}$ & $\mathcal{E}_{T}$ \\
\hline
$T=50$  & 9.73\%  & 1.90\% \\ \hline
$T=250$ & 8.86\% & 1.93\% \\ \hline
$T=500$ & 10.06\% & 3.19\% \\ \hline
\end{tabular}
\caption{Relative L$^2$ errors for the case where we 
randomly sample one or two sources with fixed width. Here, $\mathcal{E}_{\text{full}}$ denotes the average error across all time steps, and $\mathcal{E}_T$ denotes the error at final time $t = T$.}
\label{tab:double_source_errors}
\end{table}

\subsection{Triple Source with Fixed Width}
\label{sec:triple_source}
In this configuration, we consider a more structured but challenging multi-source scenario in which each input sample contains exactly three Gaussian sources with fixed spatial width. As before, the centers are drawn independently from the source subdomain $\Omega_{\text{src}} = [3,7] \times [3,7]$, and the spatial width is set to $\sigma = 0.45$ for all components. The input function is defined as:

$$
f(\mathbf{x}) = \frac{1}{\mathcal{N}} \sum_{i=1}^{3} \exp\left( -\frac{ \lVert \mathbf{x} - \mathbf{x}_i \rVert^2 }{2\sigma^2} \right),
$$

where $\mathcal{N}$ is a normalization constant. This case probes the model’s ability to learn complex source–solution mappings under compounded spatial interactions. Compared to the mixed one/two-source case, the triple-source setting results in denser and often overlapping source profiles, with higher peak magnitudes and more intricate downstream transport behavior.

To accommodate the increased complexity of the input function class in this triple-source setting, we made modest adjustments to the model architecture and training schedule. Specifically, we use a deeper network with 4 hidden layers and 256 neurons per layer, and set the learning rate decay schedule with a step size of 10,000 and decay rate of 0.97. The model is trained for $7 \times 10^5$ epochs using the same number of collocation and residual points as in the single-source case.

The results are summarized in Table-\ref{tab:triple_source_errors}. While the relative errors are slightly higher than those reported in earlier cases, as expected due to the added spatial complexity, the model still achieves strong predictive performance across all final times. Notably, the error at the final time step remains under 9\% for all three test horizons. 
% These findings suggest that even in the presence of dense source overlap and increased nonlinearity, the PI-DeepONet can effectively learn the operator mapping without requiring increased sampling density or computational budget.

\begin{table}[H]
\centering
\begin{tabular}{|l|l|l|}
\hline
\textbf{Final Time} & $\mathcal{E}_{\text{full}}$ & $\mathcal{E}_{T}$ \\
\hline
$T=50$  & 12.74\%  & 8.92\% \\ \hline
$T=250$ & 12.48\% & 9.44\% \\ \hline
$T=500$ & 14.22\% & 10.52\% \\ \hline
\end{tabular}
\caption{Relative L$^2$ errors for the case where we have triple source fixed width. Here, $\mathcal{E}_{\text{full}}$ denotes the average error across all time steps, and $\mathcal{E}_T$ denotes the error at final time $t = T$.}
\label{tab:triple_source_errors}
\end{table}

\begin{figure}[H]
    \centering

    \begin{subfigure}{0.45\textwidth}
      \stackinset{l}{0.5mm}{t}{0.5mm}{\large \color{black} \textbf{(a)}}{\fbox{\includegraphics[width=\linewidth]{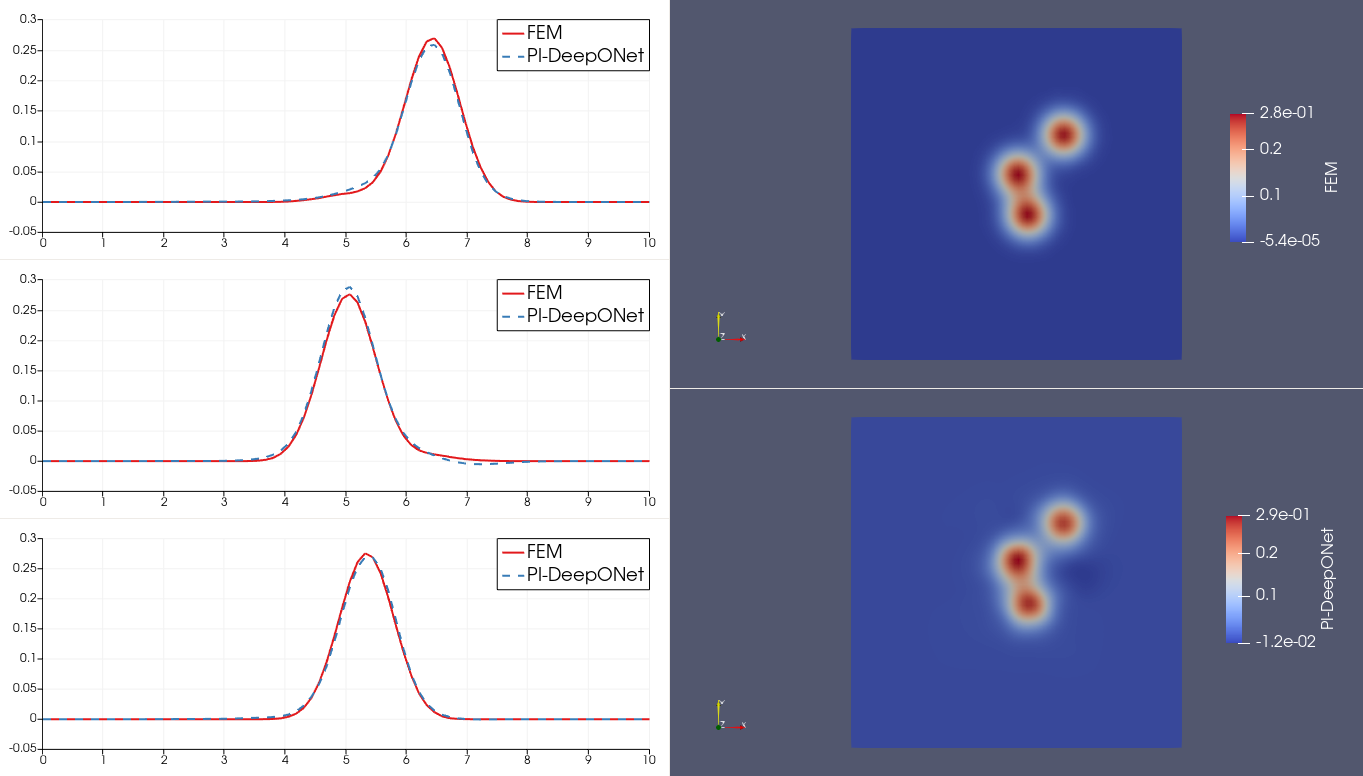}}}
    \end{subfigure} \hspace{4pt}
    \begin{subfigure}{0.45\textwidth}
      \stackinset{l}{1mm}{t}{1mm}{\large  \color{black} \textbf{(b)}}{\fbox{\includegraphics[width=\linewidth]{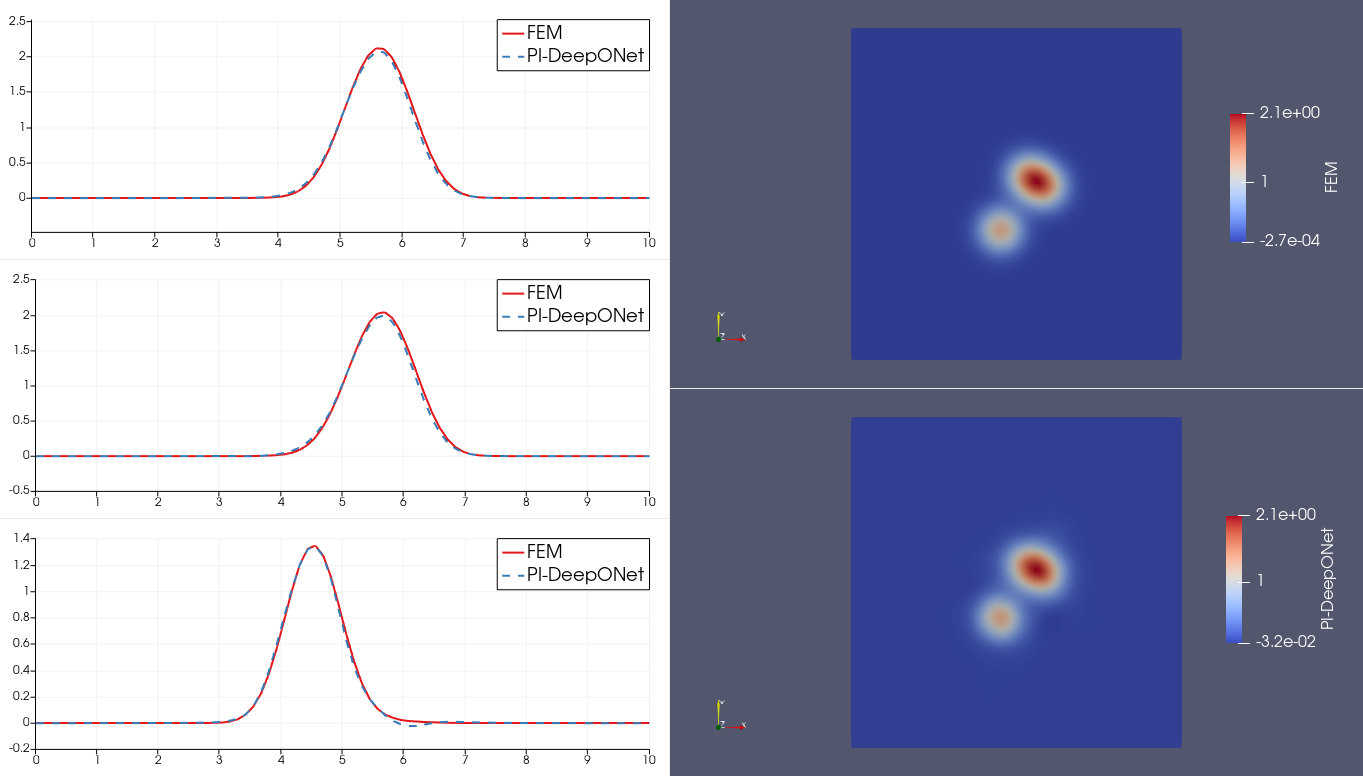}}}
    \end{subfigure}
    
    \vspace{5pt}
    
    \begin{subfigure}{0.45\textwidth}
      \stackinset{l}{1mm}{t}{1mm}{\large  \color{black} \textbf{(c)}}{\fbox{\includegraphics[width=\linewidth]{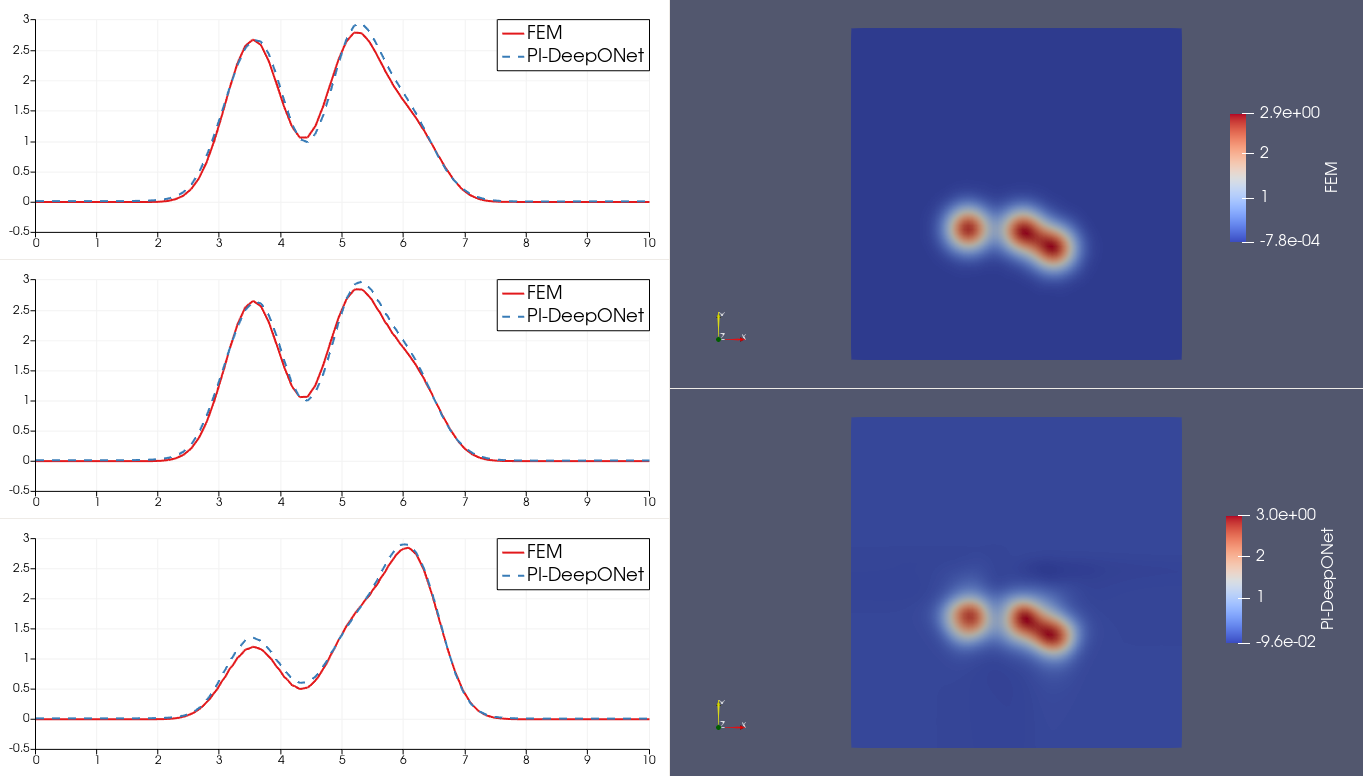}}}
    \end{subfigure}
\caption{
Comparison of PI-DeepONet and FEM solutions for three test cases at final times $T = 50$, $250$, and $500$, with each image corresponding to one of the final times, respectively. In each image:  \textbf{Left Panel}: three 1D slices passing through the $y$-centers of the three Gaussian sources.
\textbf{Right Panel}: the full spatial concentration field from FEM (top) and PI-DeepONet (bottom) is shown on the right of each sub-image. 
\textbf{(a)} $T = 50$, triple source with centers $x_0 = [(5.04,5.61),(5.35,4.36),(6.43,6.78)]$  and $\sigma = 0.45$
\textbf{(b)} $T = 250$, triple source with centers $x_0 = [(5.35,5.54),(4.52,3.91),(5.86,5.22)]$  and $\sigma = 0.45$
\textbf{(c)} $T = 500$, triple source with centers $x_0 = [(5.18,3.89),(3.55,3.95),(6.13,3.37)]$  and $\sigma = 0.45$
}
\end{figure}

\subsection{Triple Source with Fixed Centers and Varying Width}
\label{sec:triple_fixed}
As a complementary experiment to the previous section, we now consider the inverse configuration: the centers of the three Gaussian sources are held fixed, while their widths are independently varied. Each sample contains three Gaussians centered at \textit{fixed locations} and the spatial width of each Gaussian is drawn from the uniform distribution $\sigma_i \sim \mathcal{U}(0.25, 0.60)$. The resulting source term is given by:

$$
f(\mathbf{x}) = \frac{1}{\mathcal{N}} \sum_{i=1}^{3} \exp\left( -\frac{ \lVert \mathbf{x} - \mathbf{x}_i \rVert^2 }{2\sigma_i^2} \right),
$$

where $\mathcal{N}$ is a normalization constant, and the centers $\{ \mathbf{x}_i \}$ are fixed to coordinates $(3,3)$, $(4,6)$, and $(7,4)$ across all training and test instances. The network architecture and training schedule are kept identical to those used in Section 3.2.2.

As summarized in Table-\ref{tab:fixed_triple_source}, the model exhibits lower error compared to the fully variable triple-source case. Across all three final times, the relative error at the final time step drops to the 2–4\% range, suggesting that the task is more learnable when the spatial layout of the sources remains fixed. 

\begin{table}[H]
\centering
\begin{tabular}{|l|l|l|}
\hline
\textbf{Final Time} & $\mathcal{E}_{\text{full}}$ & $\mathcal{E}_{T}$ \\
\hline
$T=50$  & 9.81\%  & 2.19\% \\ \hline
$T=250$ & 9.91\% & 2.81\% \\ \hline
$T=500$ & 10.05\% & 3.12\% \\ \hline
\end{tabular}
\caption{Relative L$^2$ errors for the case where we have triple source with fixed centers and varying width. Here, $\mathcal{E}_{\text{full}}$ denotes the average error across all time steps, and $\mathcal{E}_T$ denotes the error at final time $t = T$.}
\label{tab:fixed_triple_source}
\end{table}

\begin{figure}[H]
    \centering

    \begin{subfigure}{0.45\textwidth}
      \stackinset{l}{0.5mm}{t}{0.5mm}{\large \color{black} \textbf{(a)}}{\fbox{\includegraphics[width=\linewidth]{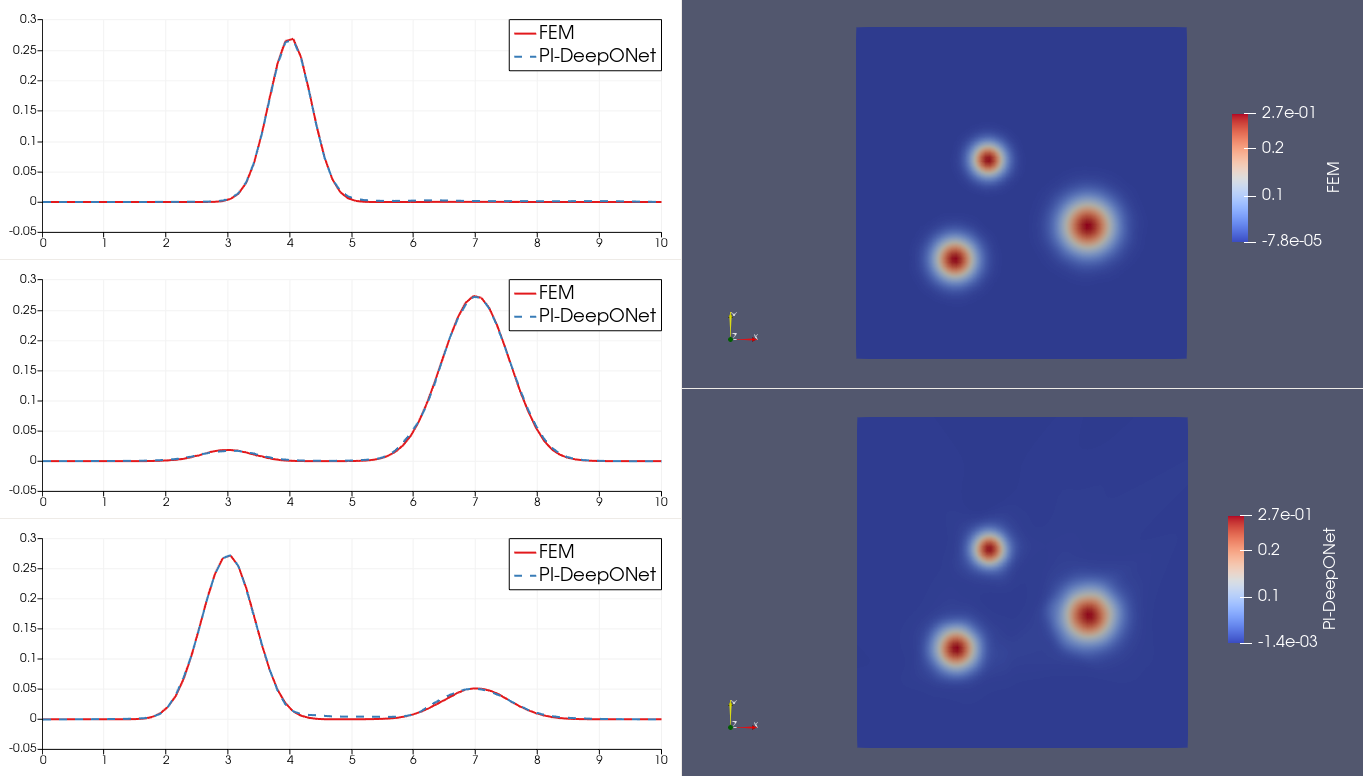}}}
    \end{subfigure} \hspace{4pt}
    \begin{subfigure}{0.45\textwidth}
      \stackinset{l}{1mm}{t}{1mm}{\large  \color{black} \textbf{(b)}}{\fbox{\includegraphics[width=\linewidth]{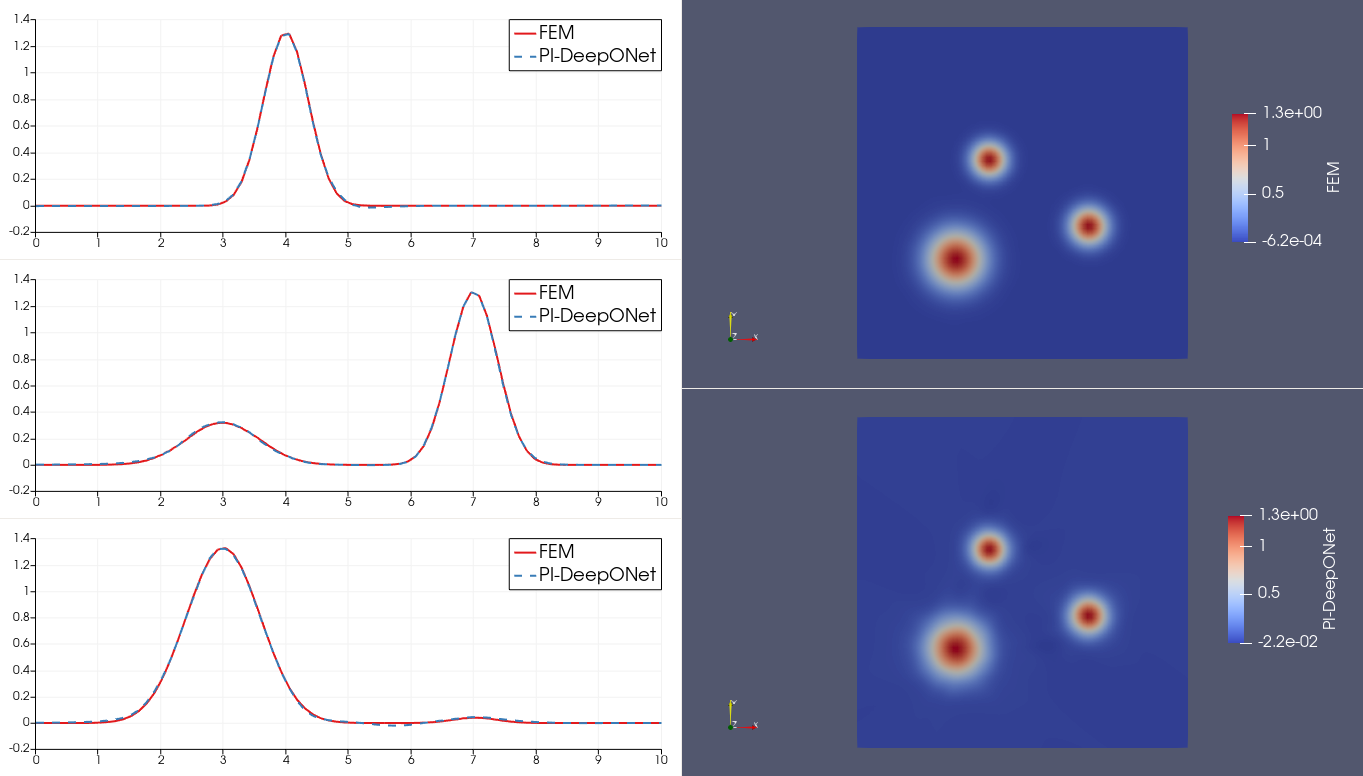}}}
    \end{subfigure}
    
    \vspace{5pt}
    
    \begin{subfigure}{0.45\textwidth}
      \stackinset{l}{1mm}{t}{1mm}{\large  \color{black} \textbf{(c)}}{\fbox{\includegraphics[width=\linewidth]{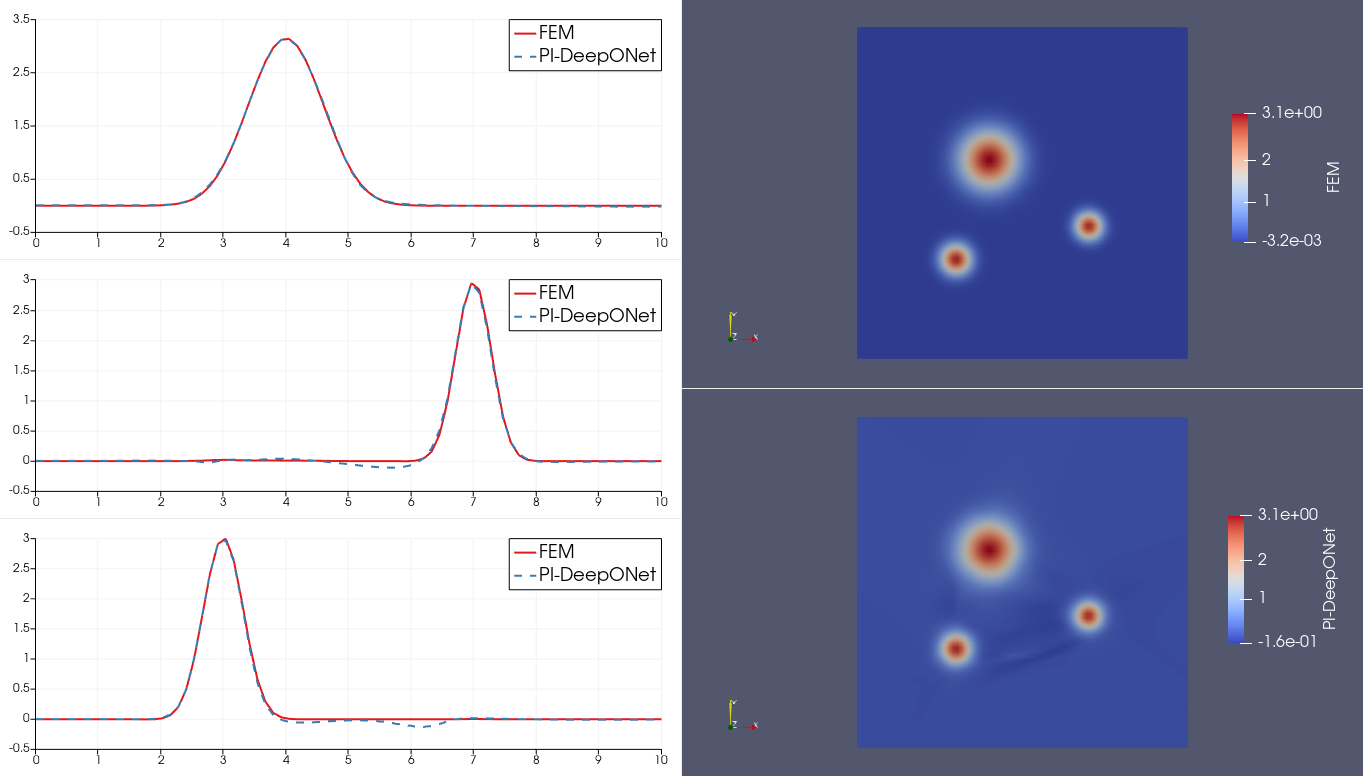}}}
    \end{subfigure}
\caption{
Comparison of PI-DeepONet and FEM solutions for three test cases at final times $T = 50$, $250$, and $500$, with each image corresponding to one of the final times, respectively. In each image:
Left Panel: three 1D slices passing through the $y$-centers of the three Gaussian sources.
Right Panel: the full spatial concentration field from FEM (top) and PI-DeepONet (bottom) is shown on the right of each sub-image.
(a) $T = 50$, triple source with centers $x_0 = [(3,3),(4,6),(7,4)]$ and $\sigma = (0.43,\ 0.34,\ 0.55)$, respectively.
(b) $T = 250$, triple source with centers $x_0 = [(3,3),(4,6),(7,4)]$ and $\sigma = (0.59,\ 0.36,\ 0.38)$, respectively.
(c) $T = 500$, triple source with centers $x_0 = [(3,3),(4,6),(7,4)]$ and $\sigma = (0.46,\ 0.46,\ 0.37)$, respectively.
}
\label{fig:fixed_triple_source}
\end{figure}

Figure-\ref{fig:fixed_triple_source} presents representative model outputs for final times $T = 50$, $250$, and $500$. The concentration fields exhibit close agreement with the FEM ground truth, including accurate peak heights and spatial spread. In the last row, some minor visual artifacts are present in the PI-DeepONet prediction, likely due to subtle convergence irregularities, but the overall shape and magnitude remain well-preserved.

\subsection{Performance in a Higher-Péclet Regime with Increased Permeability}
Throughout our previous experiments, the intrinsic permeability was fixed at $K = 10^{-9}$, corresponding to a maximum Péclet number of approximately $0.48$, placing the system in a diffusion-dominated regime. To assess the performance of the PI-DeepONet framework under more convection-dominated conditions, we increased the permeability by one order of magnitude to $K = 10^{-8}$, which raises the maximum Péclet number to about $4.8$.

We repeated the experiments described in Sections \ref{sec:single_source} and \ref{sec:triple_fixed} under this new setting. The results are summarized in Table-\ref{tab:high_Perm}.

\begin{table}[H]
\centering
\begin{tabular}{|>{\raggedright\arraybackslash}p{7cm}|c|c|}
\hline
\textbf{Experiment (T=500)} & $\boldsymbol{\mathcal{E}_{\text{full}}}$ & $\boldsymbol{\mathcal{E}_T}$ \\
\hline
Single Source with Varying Center and Width & 10.84 & 5.70 \\ \hline
Triple Source with Fixed Centers and Varying Width & 10.56 & 4.04 \\
\hline
\end{tabular}
\caption{Relative L$^2$ errors for two difference experiments with permeability $K=10^{-8}$. $\mathcal{E}_{\text{full}}$ denotes the average error across all time steps, and $\mathcal{E}_T$ denotes the error at final time $t = 500$.}
\label{tab:high_Perm}
\end{table}

\begin{figure}[H]
    \centering
    \begin{subfigure}{0.85\textwidth}
      \stackinset{l}{0.5mm}{t}{0.5mm}{\large \color{black} \textbf{(a)}}{\fbox{\includegraphics[width=\linewidth]{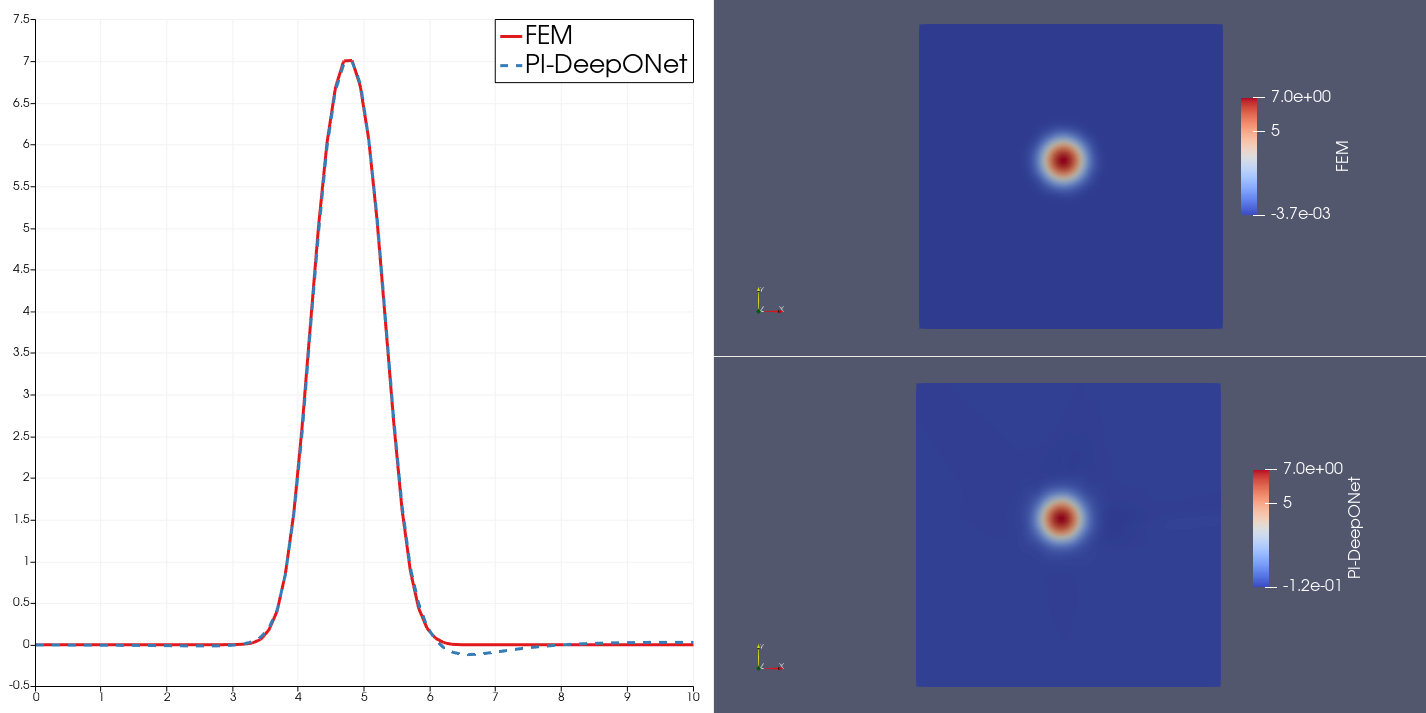}}}
    \end{subfigure} \\ \vspace{10pt}
    \begin{subfigure}{0.85\textwidth}
      \stackinset{l}{1mm}{t}{1mm}{\large  \color{black} \textbf{(b)}}{\fbox{\includegraphics[width=\linewidth]{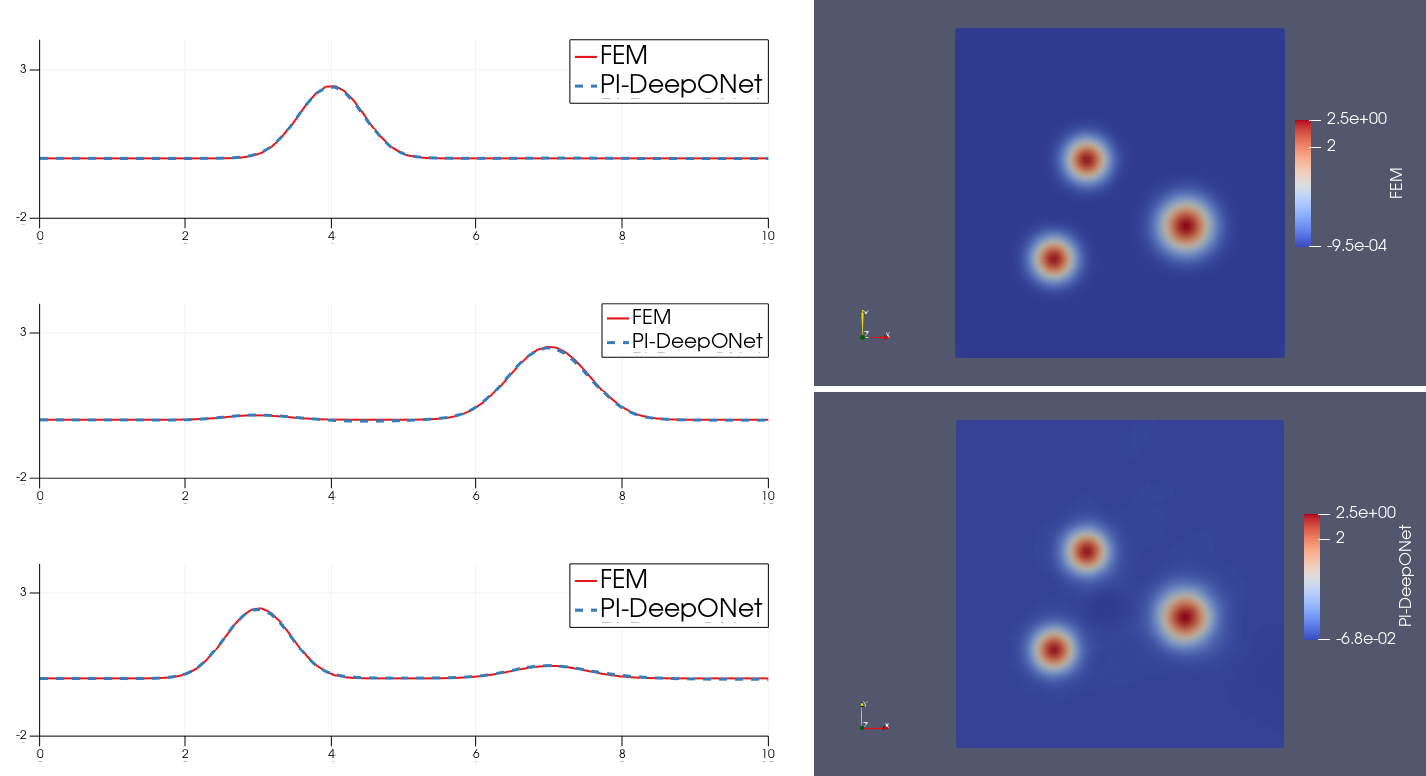}}}
    \end{subfigure}
\caption{
Comparison of PI-DeepONet and FEM solutions for two test cases at final time $T=500$ with permeability $K=10^{-8}$. In each image:
Left Panel: three 1D slice(s) passing through the $y$-center(s) of the Gaussian source(s).
Right Panel: the full spatial concentration field from FEM (top) and PI-DeepONet (bottom) is shown on the right of each sub-image.
\textbf{(a)} Single source with center $x_0 = (4.75,5.53)$ and $\sigma = 0.44$.
\textbf{(b)} Triple source with centers $x_0 = [(3,3),(4,6),(7,4)]$ and $\sigma = (0.42,\ 0.42,\ 0.52)$, respectively.
}
\label{fig:higher_perm}
\end{figure}

In both experiments, the model maintains accuracy comparable to that observed in the low-convective regime. Figure-\ref{fig:higher_perm} shows representative test cases for each setting. Due to the increased convective transport, peak concentration values are reduced: from approximately $12 \text{mmol}$ to $7 \text{mmol}$ in the single-source case, and from about $3.1$ to $2.5$ in the triple-source case, as the injected solute is advected away from the source region more rapidly.

\subsection{Effect of Residual Point Sampling Strategy}

To better understand the role of our spatial residual point sampling strategy, we conducted an ablation study comparing the standard hybrid sampling scheme to a purely random baseline. In our standard setup, residual collocation points are composed of a structured polar grid, using $n_r = 30$ and $n_\theta = 30$, centered at the Gaussian source, combined with $n_{\mathrm{rand}} = 300$ additional points randomly distributed over the domain. This totals to 1200 and 3000 residual points per instance for single source and triple source cases respectively.

To isolate the effect of the structured sampling, we reran each experiment with the same total number of residual points but removed the polar grid entirely, setting $n_r = n_\theta = 0$ and allocating all points to random sampling. This comparison was repeated for three representative experiments in Sec-\ref{sec:single_source},\ref{sec:triple_source} and \ref{sec:triple_fixed}. All other model and training parameters were kept constant, and results were reported for the final-time case $T = 500$.

The results are summarized in Table-\ref{tab:sampling_ablation}. Across all cases, replacing the dense grid with random points leads to significantly worse performance. In particular, the final-time error $\mathcal{E}_T$ increases by a significant margin across all experiments. This confirms that dense sampling around the source locations is critical for guiding the model to resolve steep gradients and local transport effects. 

\begin{table}[H]
\centering
\begin{tabular}{|l|c|c|c|c|c|}
\hline
\textbf{Experiment} & $n_r$ & $n_\theta$ & $n_{\mathrm{rand}}$ & $\mathcal{E}_{\text{full}}$ & $\mathcal{E}_{T}$ \\
\hline
Single Source with Varying Center and Width 
& 30 & 30 & 300 & 10.70\% & 4.90\% \\
\cline{2-6}
& 0  & 0  & 1200 & 14.76\% & 14.10\% \\
\hline
Triple Source with Fixed Width             
& 30 & 30 & 300  & 12.59\% & 8.92\% \\
\cline{2-6}
& 0  & 0  & 3000 & 31.11\% & 29.10\% \\
\hline
Triple Source with Fixed Centers and Varying Width 
& 30 & 30 & 300  & 10.33\% & 3.73\% \\
\cline{2-6}
& 0  & 0  & 3000 & 36.53\% & 38.32\% \\
\hline
\end{tabular}
\caption{Ablation study evaluating the impact of residual point sampling strategy on model performance. For each experiment, the top row shows results using hybrid sampling (structured polar grid around the source plus random points), while the bottom row uses only randomly sampled residual points. The total number of residual points is kept fixed across both configurations.}
\label{tab:sampling_ablation}
\end{table}

\subsection{Runtime Comparison: FEM vs. PI-DeepONet}

To evaluate the computational benefits of the learned PI-DeepONet model, we compare its inference time against the traditional FEM solver across four representative test cases at final time $T=500$. For a fair comparison, all file I/O operations on the FEM side were disabled, and timings were averaged over 30 test samples per experiment. As shown in Table~\ref{tab:speedup_comparison}, the PI-DeepONet model consistently achieves two to three orders of magnitude speed-up over the FEM baseline. The speed-up becomes especially pronounced in the more complex triple-source cases, reaching nearly $\times 300$ in multi-source setting.

\begin{table}[H]
\centering
\begin{tabular}{|>{\raggedright\arraybackslash}p{5.8cm}|c|c|c|c|}
\hline
\textbf{Experiment (T=500)} & \textbf{FEM (s)} & \textbf{Training Time (min)} & \textbf{Model (s)} & \textbf{Speed-up} \\
\hline
Single Source with Varying Center and Width & 5.265 & 84.53 & 0.038 & 138.399 \\ \hline
Random Sampling of One or Two Sources with Fixed Width & 6.564 & 83.72 & 0.039 & 168.713 \\ \hline
Triple Source with Fixed Width & 9.081 & 102.13 & 0.031 & 289.739 \\ \hline
Triple Source with Fixed Centers and Varying Width & 11.809 & 102.04 & 0.036 & 328.153 \\
\hline
\end{tabular}
\caption{Average execution times (in seconds) for FEM and PI-DeepONet over 30 test cases per experiment.}
\label{tab:speedup_comparison}
\end{table}

\newpage
\section{Discussion and Future Directions}
\label{sec:discussion}
% \erdi{
% \begin{itemize}
%     \item Continual learning aspect, moving from single to multiple gaussian case by resuming training.
%     \item Practical use cases where this framework might be useful.
%     \item Low fidelity quick approximation. 
%     \item Ready to use, extensively documented library for coupling fem and PI-DeepoOnet. This is not an easy business(making Petsc and Jax and Pytorch together )
% \end{itemize}
% }

Our results demonstrate that the proposed FEM–PI-DeepONet pipeline can accurately learn the map $f \mapsto c$ even in the presence of sharp, spatially localized Gaussian sources and a coupled Darcy–Transport system. Unlike standard DeepONet demonstrations, which often focus on isolated PDEs with smooth input functions drawn from Gaussian random fields or low-frequency bases, our setup involves a fully physics-informed training objective, where the target equation is a time-dependent convection–diffusion system governed by a precomputed Darcy velocity field. While the Gaussian source functions used here are smooth, they are highly localized and sharply peaked, leading to steep gradients and requiring fine spatial resolution. Despite these challenges, we show that accurate learning is possible with as few as 2000 training functions and moderate computational resources, a consumer laptop with an Intel i7-11800H CPU and NVIDIA RTX 3080 GPU was used for all experiments.

The resulting model achieves up to $\times 300$ speed-up compared to full FEM solvers while maintaining strong agreement in final-time solute concentration profiles. This acceleration is especially valuable for settings that require repeated queries, such as uncertainty quantification, inverse design, or real-time control, where traditional FEM approaches become prohibitively expensive. 
A key advantage of our approach is its modularity: the FEM solver is used solely to compute the velocity field, while the operator network independently learns the mapping from the source function to concentration. The meshing, Darcy solver, and PI-DeepONet architecture are fully decoupled, allowing one to easily swap out components. For example, one can retrain the operator network using a different residual grid or with updated flow inputs, without modifying the FEM solver infrastructure. Since the velocity field is precomputed and mapped to the collocation domain before training, the FEM and PI-DeepONet components remain modular and independently configurable.

Our model performs well within the convective regime considered in this work; however, pushing to more strongly advection-dominated flows becomes increasingly difficult due to the combined effect of high Péclet numbers and sharply localized Gaussian sources. Moreover, while the Péclet number can be increased further by raising $K$, the combination of higher advection and sharply localized Gaussians leads to steep gradients that are difficult even for FEM to resolve, making such scenarios especially demanding for PI-DeepONet. Another important consideration is that, although trunk network collocation points are adaptively sampled, the branch network sensor grid is fixed by design. Recall that, in our experiments, the Gaussian width is drawn from $\sigma \sim U(0.25, 0.60)$, which corresponds to an average coverage of only about $3.7\%$ of the domain area. Using a fixed $30\times30$ branch grid is adequate in this range, but pushing $\sigma$ to smaller values would require a finer grid to capture the source accurately. This, in turn, would substantially increase the dimensionality of the branch network input layer, and hence the total number of trainable parameters, without necessarily adding proportional expressive power. This sensitivity to sensor resolution is a general limitation of DeepONet-based models and several recent works attempt to address this limitation \cite{tretiakov2025setonet,bahmani2025resolution,ingebrand2025basis}.

From a physical modeling standpoint, an important limitation arises when the source term in the transport equation becomes extremely weak. Recall that the key parameter governing this term is the source coefficient $\beta_2$, which represents the rate of solute mass injection per unit volume. Physically, $\beta_2$ is the product of the volumetric infusion rate, which we already defined as $\beta_1$, and the solute concentration of the infused fluid $c_{\inf}$, so that $\beta_2 = \beta_1 c_{\inf}$. When $c_{\inf}$ is small, as is typically the case in realistic applications, the effective source term $\beta_2 f$ becomes negligible. As a result, the solute concentration field is primarily governed by advection, and the mapping $f \mapsto c$ becomes nearly unidentifiable. To retain a learnable signal in this study, we used $c_{\inf} = 1/4$, but this is significantly higher than what is physically realistic. In future work, we plan to address this limitation by learning the full mapping $(\mathbf{v}, f) \mapsto c$, which more accurately reflects the underlying physics and may enable learning even when the source signal is weak. This extension introduces additional challenges related to velocity field representation and input dimensionality, but it is a necessary step toward modeling more realistic systems.

While the model performs very well at final times, its accuracy is more limited during earlier stages of the simulation. For example, in the case with final time $T = 500$, the final-time relative $L^2$ error is approximately 5\%, while the average error across all time steps is around 10\%. 
This higher error is mainly due to the model not converging well at early times, which is expected for two reasons. First, the framework lacks an explicit notion of temporal causality which train over successive time steps to enforce sequential consistency, with significantly higher computational cost \cite{wang2024respecting,krishnapriyan2021characterizing}. Second, the model is subject to spectral bias: the localized Gaussian sources typically span only a small portion of the domain, which means that early-time signals are weak and spatially confined, making them harder to resolve. Despite these challenges, the model successfully captures the full concentration profile and achieves high accuracy at the final time, which is the primary target of interest in this study.

The training process also reveals that progressive learning, from single to multi-source configurations, improves stability and generalization. Although not explored in detail in this paper, initial experiments indicate that curriculum-based training strategies may be highly beneficial for operator learning in multiscale PDE systems. This is consistent with broader trends in curriculum learning literature, though applying such ideas to DeepONet remains challenging due to catastrophic forgetting and a lack of robust scheduling strategies \cite{wang2022and,dekhovich2025ipinns,howard2024multifidelity}.
.

In our pipeline, model training is performed in JAX \cite{jax2018github}, while data loading and batching are handled using PyTorch utilities \cite{paszke2019pytorch} to enable efficient preprocessing and integration with standard ML workflows. All FEM-related computations are carried out using Firedrake \cite{rathgeber2016firedrake}. We should note that a substantial software engineering effort is required to integrate these tools into a cohesive and performant workflow. This combination is powerful but technically demanding, often prone to subtle bugs and compatibility issues. Our implementation abstracts these complexities away and will be released in a Dockerized environment, allowing users to reproduce our experiments with no installation burden. We believe this kind of reproducibility and accessibility is essential for pushing the field forward. 
This effort aligns with a growing body of work aiming to bridge the gap between scientific computing and machine learning through differentiable programming and unified toolchains \cite{bezgin2023jax, jaxmd2020, xue2023jax}. Projects like JAX-Fluids \cite{bezgin2023jax}, JAX-MD \cite{jaxmd2020}, and JAX-FEM \cite{xue2023jax} are some examples of this trend trying to incorporate differentiable simulations with modern ML libraries. Our contribution follows this trajectory, demonstrating that tightly coupled PDE solvers and neural networks can be utilized in a unified, reproducible manner in ML-driven scientific scientific computing.

\section*{Acknowledgments}

This project was completed with support from the U.S. Department of Energy, Advanced Scientific Computing Research
program, under the "Uncertainty Quantification for Multifidelity Operator Learning (MOLUcQ)" project (Project No. 81739).
The research was also funded by the Simons Foundation through a grant made to Spelman College. Pacific Northwest National Laboratory (PNNL) is a multi-program national laboratory operated for the U.S. Department of
Energy (DOE) by Battelle Memorial Institute under Contract No. DE-AC05-76RL01830.

\clearpage
\newpage
\bibliographystyle{unsrt}
\bibliography{ref}

\end{document}